\documentclass{article} %
\usepackage{iclr2024_conference,times}

\iclrfinalcopy

\usepackage{hyperref}
\usepackage{url}
\usepackage{CJKutf8}
\usepackage[final,nopatch=footnote]{microtype}

\usepackage{amsmath,amsfonts,amsthm,amssymb,bm}

\def\eqref#1{equation~\ref{#1}}

\def\1{\bm{1}}

\DeclareMathAlphabet{\mathsfit}{\encodingdefault}{\sfdefault}{m}{sl}
\SetMathAlphabet{\mathsfit}{bold}{\encodingdefault}{\sfdefault}{bx}{n}

\newcommand{\R}{\mathbb{R}} %

\theoremstyle{definition}

\theoremstyle{definition}

\usepackage{style}

\usepackage{algpseudocode,algorithmicx}
\usepackage{wrapfig}

\definecolor{orange}{rgb}{1,0.5,0}
\definecolor{mdred}{rgb}{0.7,0,0}
\definecolor{mdgreen}{rgb}{0.05,0.6,0.05}
\definecolor{mdblue}{rgb}{0,0,0.7}
\definecolor{dkblue}{rgb}{0,0,0.5}
\definecolor{dkgray}{rgb}{0.3,0.3,0.3}
\definecolor{slate}{rgb}{0.25,0.25,0.4}
\definecolor{gray}{rgb}{0.5,0.5,0.5}
\definecolor{ltgray}{rgb}{0.7,0.7,0.7}
\definecolor{purple}{rgb}{0.7,0,1.0}
\definecolor{lavender}{rgb}{0.65,0.55,1.0}
\definecolor{real}{HTML}{3498DB}
\definecolor{cf}{HTML}{DB7734}

\newcommand{\papercomment}[3]{\ensuretext{\textcolor{#3}{[#1 #2]}}}
\renewcommand{\papercomment}[3]{}  %

\newcommand{\token}[1]{``\texttt{#1}''\xspace}
\newcommand{\chin}[1]{\begin{CJK*}{UTF8}{gbsn}#1\end{CJK*}}
\newcommand{\dom}{\star}
\newcommand{\nodom}{\circ}

\title{
The Semantic Hub Hypothesis: \\ Language Models Share  Semantic Representations Across  Languages and Modalities
}

\author{Zhaofeng Wu$^\text{\Cancer}$ \quad Xinyan Velocity Yu$^\text{\Gemini}$ \quad Dani Yogatama$^\text{\Gemini}$ \quad Jiasen Lu$\rotatebox[y=0.25cm]{90}{\textsuperscript{\Gemini}}$ \quad Yoon Kim$^\text{\Cancer}$ \\
$^\text{\Cancer}$MIT \quad
$^\text{\Gemini}$University of Southern California \quad
$\rotatebox[y=0.25cm]{90}{\textsuperscript{\Gemini}}$Allen Institute for AI \\
\texttt{zfw@csail.mit.edu}
}

\begin{document}

\maketitle
\vspace{-2mm}
\begin{abstract}
\vspace{-2mm}
Modern language models can process inputs across diverse languages and modalities.
We hypothesize that models acquire this capability through learning a \emph{shared representation space} across heterogeneous data types (e.g., different languages and modalities), which places semantically similar inputs near one another, even if they are from different modalities/languages. We term this the \emph{semantic hub hypothesis}, following the hub-and-spoke model from neuroscience \citep{patterson2007you} which posits that semantic knowledge in the human brain is organized through a transmodal semantic ``hub'' which integrates information from various modality-specific ``spokes'' regions.
We first show that model representations for semantically equivalent inputs in different languages are similar in the intermediate layers,  and that this space can be interpreted using the model's dominant pretraining language via the logit lens.
This tendency extends to other data types, including arithmetic expressions, code, and visual/audio inputs.
Interventions in the shared representation space in one data type also predictably affect model outputs in other data types,
suggesting that this shared representations space is not simply a vestigial byproduct of large-scale training on broad data, but something that is actively utilized by the model during input processing.
\end{abstract}

\begin{figure}[h!]
    \centering
    \includegraphics[width=\textwidth]{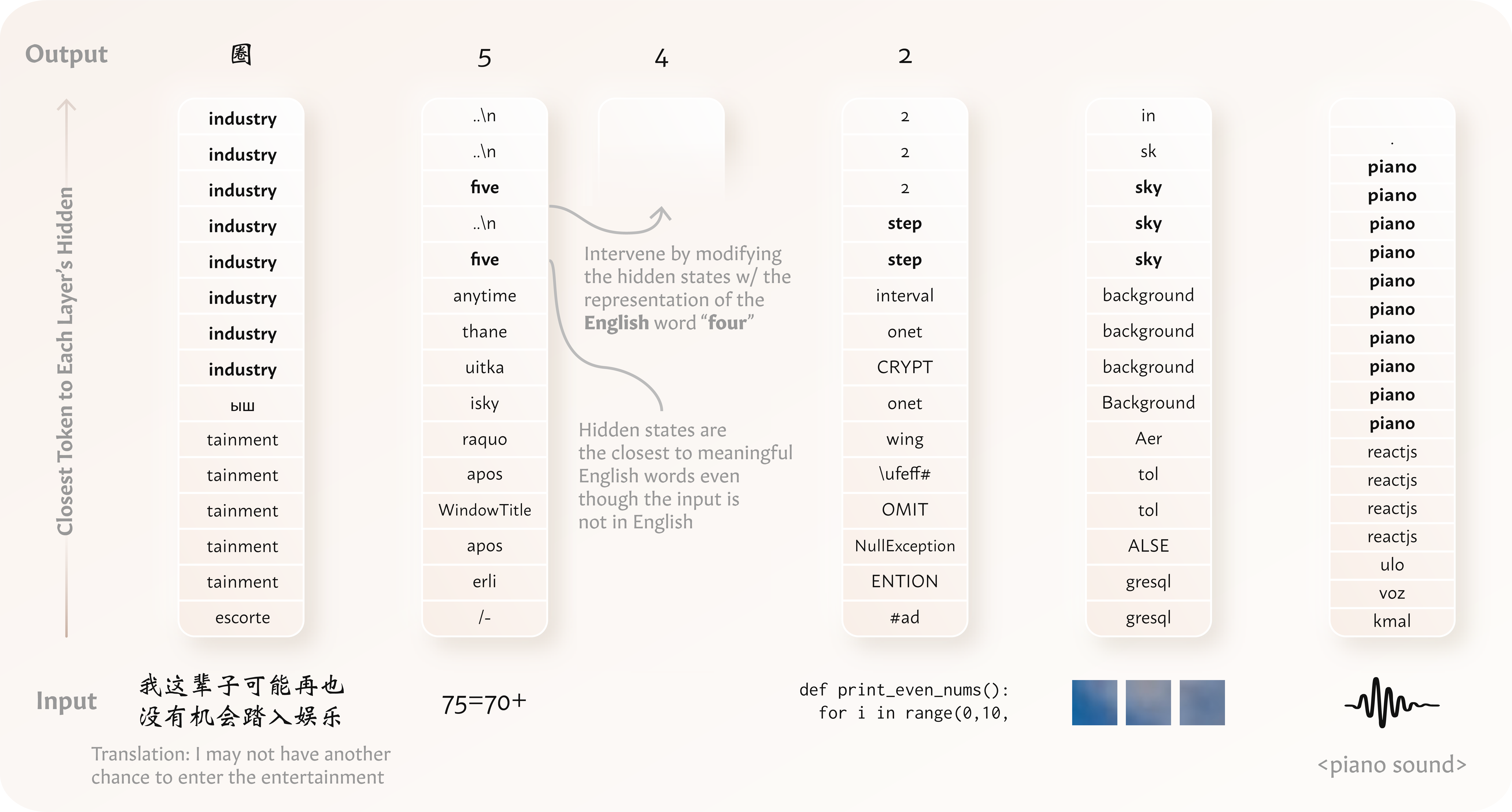}
    \caption{Examples of the semantic hub effect across input data types. For every other layer, we show the closest output token to the hidden state based on the logit lens. Llama-3's hidden states are often closest to English tokens when processing Chinese texts, arithmetic expressions, and code, in a semantically corresponding way. LLaVA, a vision-language model, and SALMONN, an audio-language model, have similar behavior when processing images/audio. As shown for the arithmetic expression example, models can be intervened cross-lingually or cross-modally, such as using English even though the input is non-English, and be steered towards corresponding effects. Boldface is only for emphasis.}
    \label{fig:overview}
    \vspace{-6mm}
\end{figure}

\vspace{-2mm}
\section{Introduction}
\vspace{-2mm}
Modern language and multimodal models (LMs)\footnote{Hereafter, we use the term ``language model'' loosely and also consider multimodal language models that process additional data modalities, since such models are commonly trained on top of a text LM backbone.}  are capable of processing heterogeneous data types: text in different languages, non-linguistic inputs such as code and math expressions, and even other modalities such as images and sound. How do LMs process these distinct data types with a single set of parameters? One strategy might be to learn specialized subspaces for each data type that are only employed when processing it. In many cases, however, data types that are surface-distinct share underlying semantic concepts. This is most obvious for sentences in different languages with the same meaning; but such shared concepts are present across other data types, e.g., between an image and its caption, or a piece of code and its natural language description.
The human brain, for example, is believed to have a transmodal ``semantic hub''~\citep{patterson2007you,ralph2017neural} located in the anterior temporal lobe that integrates and stores semantic information from various modality-specific ``spokes'' (e.g., visual/auditory cortices).
A model, leveraging the structural commonalities across data types, could similarly project their surface forms into a \emph{shared} representation space, perform computations in it, and then project back out into surface forms when needed.

To what extent is this idealized strategy adopted by actual models? \citet{wendler-etal-2024-llamas} find that on simple synthetic tasks, Llama-2~\citep{touvron2023llama2} maps various input languages into a shared ``English space'' before projecting back out into another language, hinting that it leverages this shared representation scheme to an extent, at least for different languages. We show that this is a much more general phenomenon: \textbf{when a model processes inputs from multiple data types, there is a shared representation space, and this space is scaffolded by the LM's inherently dominant language (usually English)}. By scaffolded, we mean that the shared space can be interpreted to an extent in the dominant data type via the logit lens \citep{logit-lens}. Following the cognitive science nomenclature, we call this shared representation space the LM's ``semantic hub.''

We first show that LMs represent semantically similar inputs from distinct data types (across languages, or between natural language and arithmetic expressions, code, formal semantic structures, and multimodal inputs) to be close to one another in intermediate LM layers.
We further show that we can interpret these hidden representations to an extent using the LM's dominant data type---e.g., when processing a Chinese input, an English-dominant LM ``thinks'' in English before projecting back out to a Chinese space.
Finally, we perform intervention experiments showing that intervening in the shared representation space using the LM's dominant data type, predictably affects model output when processing other data types; that is, the shared representation space (and the processing of these representations through subsequent layers) is not a vestigial byproduct of the model's being trained on (say) English-dominant text, but causally impacts model behavior.

Our work is complementary and distinct from prior work which finds structural similarities between the representation spaces of models trained (usually independently) on different data types, such as those showing that text representations from text-only LMs can be aligned, via a transformation, to vision/audio representations of modality-specific models \citepia{ilharco2020probing,merullo2022linearly,li2023implications,ngo2024language,huh2024platonic}, and the literature on cross-lingual word embedding alignment~\citepia{mikolov2013exploiting,artetxe-etal-2017-learning,conneau2017word,schuster-etal-2019-cross}. We instead show that an LM trained on multiple data types represents and processes them in a shared space \emph{without} requiring explicit alignment transformation. We hope our findings shed light on ways to more easily interpret the mechanisms of current models and motivate future work aimed at better controlling models using these insights.\footnote{We release our code at \url{https://github.com/ZhaofengWu/semantic-hub}.}
\vspace{-2mm}
\section{The Semantic Hub Hypothesis} \label{sec:hypothesis}
\vspace{-2mm}
Let $\mathcal{X}_z$ be the domain of some data type $z\in\mathcal{Z}$ where $\mathcal{Z}$ is the set of model-supported data types. E.g., for languages, $\mathcal{X}_{\text{Chinese}}$ could be all Chinese tokens, while for images $\mathcal{X}_{\text{Image}} = [0,255]^{w \times h \times 3}$ could be the RGB values for an $w \times h$-sized image patch.
Consider a function $M_z: \mathcal{X}_z^\ast \to \mathcal{S}_z$ mapping an input sequence into a semantic representation space $\mathcal{S}_z$ (i.e., the ``hub''), and a verbalization function $V_z: \mathcal{S}_z \to \mathcal{X}_z^\ast$. %
Given an input prefix $w_{1:t}^z\in\mathcal{X}_z^\ast$ of length $t$ where  $w_i^z \in \mathcal{X}_z$, a sensible (implementation-agnostic) way to continue the sequence is to first encode the input into a modality-agnostic representation $m^{in}=M_z(w_{1:t})$, reason and formulate a representation of possible futures to obtain $m^{out}\in\mathcal{S}_z$, and finally verbalize it via $V_z(m^{out})$. 

\begin{wrapfigure}{R}{8cm}
\vspace{-4mm}
    \begin{minipage}{0.57\textwidth}
    \centering
    \includegraphics[width=\textwidth]{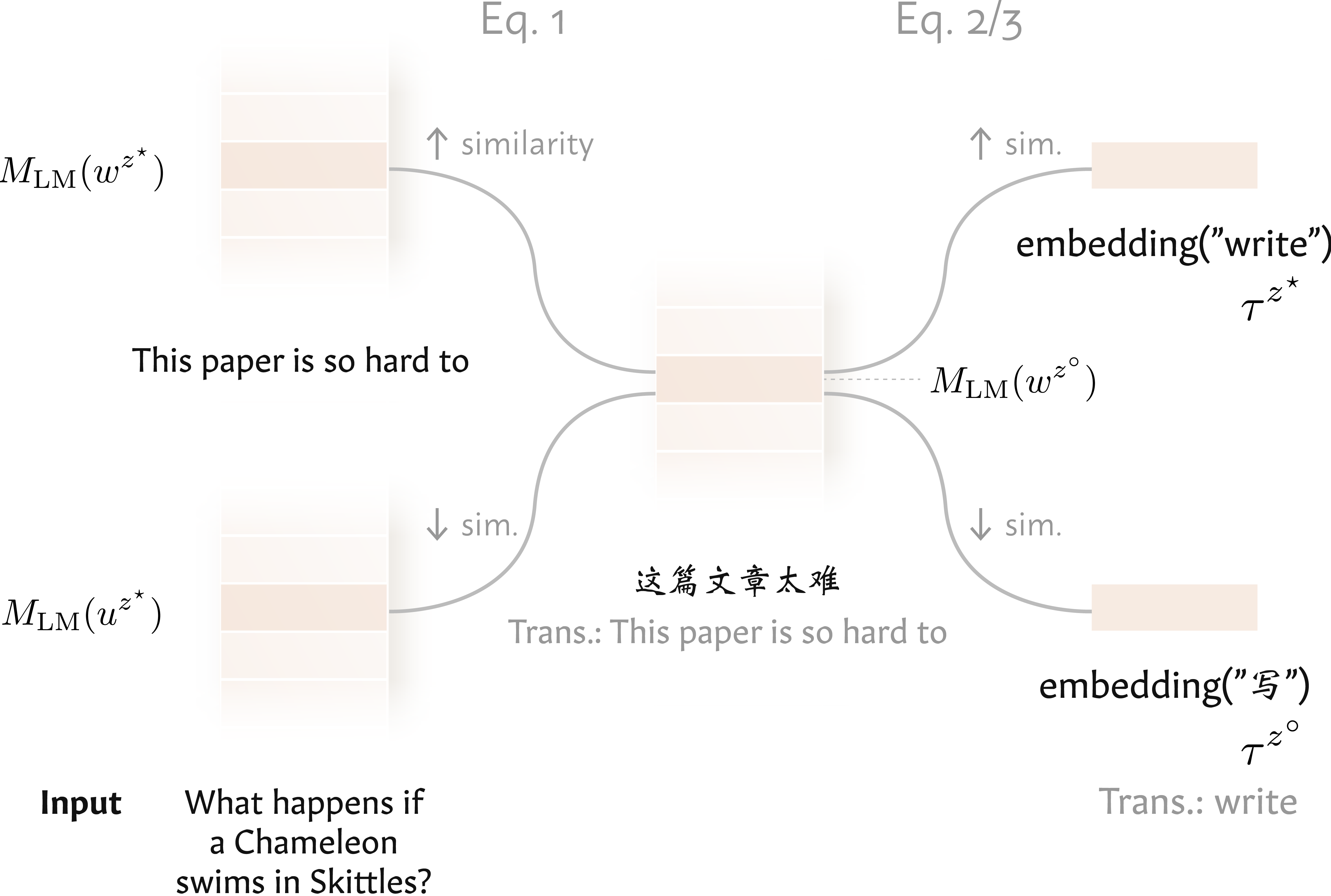}
    \caption{An illustration of our hypothesis, where semantically equivalent inputs (across data types) have similar representations, and this representation is close to the continuation token in the dominant data type. Here, $z^\dom$ is English and $z^\nodom$ is Chinese.}
    \label{fig:method}
    \end{minipage}
    \vspace{-5mm}
\end{wrapfigure}

An LM parameterizes a similar process: 
it uses $M_{\text{LM}}$ to map various input data types into 
a representation space $\mathcal{S}_\text{LM} \subseteq \mathbb{R}^d$ (early layers), performs computations in the space (middle layers), and verbalizes the output via $V_{\text{LM}}$ (end layers and the LM head).
However, it is unknown as to how different data types are structured in the representation space.
For example, one possibility is that the LM partitions $\R^d$ into disjoint subspaces for each data type and processes them separately.
We instead hypothesize that LMs, through training,  learn to represent and process different data types in a \emph{shared} representation space that functions as a modality-agnostic ``semantic hub.''
That is, semantically similar inputs $w_{1:t}^{z_1}$ and $w_{1:t'}^{z_2}$ from distinct data types---for example texts in different languages that are mutual translations---are similarly mapped in $\mathcal{S}_{\text{LM}}$; informally, $M_{\text{LM}}(w_{1:t}^{z_1})\approx M_{\text{LM}}(w_{1:t'}^{z_2})$.
However, absolute similarity measures (i.e., $\operatorname{sim}(M_{\text{LM}}(w_{1:t}^{z_1}), M_{\text{LM}}(w_{1:t'}^{z_2}))$) are generally difficult and unintuitive to interpret in high dimensional spaces.\footnote{See for example \citet{beyer1999}. Most prior work on probing also implicitly uses relative similarity measures since the similarity scores are normalized over a finite label set.} We thus focus on relative similarity measures, taking a semantically unrelated sequence $u_{1:t'}^{z_2}$, and evaluating whether $w_{1:t}^{z_1}$ is closer to $w_{1:t'}^{z_2}$ than $u_{1:t''}^{z_2}$, as illustrated on the left of Figure~\ref{fig:method}. Formally:
\begin{align} \label{eq:rep-sim-enc}
    \operatorname{sim}\left(M_{\text{LM}}(w_{1:t}^{z_1}), M_{\text{LM}}(w_{1:t'}^{z_2})\right) > \operatorname{sim}\left(M_{\text{LM}}(w_{1:t}^{z_1}), M_{\text{LM}}(u_{1:t''}^{z_2})\right).
\end{align}

Moreover, when the LM has a \emph{dominant data type} $z^\dom$ in training (e.g., English for Llama-2), we hypothesize that this shared representation space is ``anchored'' by tokens in $z^\dom$ (denoted as $\tau^{z^\dom}$),
thereby allowing the semantic hub to be interpretable.
We focus on anchor tokens that represent a continuation of the input, which autoregressive LMs are trained to model.
Informally, $M_{\text{LM}}(w_{1:t}^{z})$ is close to the embedding of token $\tau^{z^\dom}$, which represents a corresponding continuation in $z^\dom$, more than some $\tau^{z^\nodom}$ in a non-dominant data type $z^\nodom$, even when $z^\nodom$ is the input data type.
In Figure~\ref{fig:method}, e.g., with an English-dominant LM, its encoding of the Chinese prefix $w_{1:t}^{z^\nodom}=$\token{\chin{这篇论文太难}} (trans. \token{This paper is so hard to}) should be closer to the representation of the continuation word in English $\tau^{z^\dom}=$\token{write} than its Chinese translation $\tau^{z^\nodom}=$\token{\chin{写}}.
Formally:
\begin{align} \label{eq:dom-readoutability-dec}
    \operatorname{sim}\left(M_{\text{LM}}(w_{1:t}^{z^\nodom}), \operatorname{emb}(\tau^{z^\dom})\right) > \operatorname{sim}\left(M_{\text{LM}}(w_{1:t}^{z^\nodom}), \operatorname{emb}(\tau^{z^\nodom})\right).
\end{align}

\vspace{-2mm}
\subsection{Method: Testing the Semantic Hub Hypothesis} 
\vspace{-2mm}\label{sec:method-new}

We test the semantic hub hypothesis for LMs by considering pairs of distinct data types, the dominant one $z^\dom$  and a non-dominant one $z^\nodom$, different for each experiment. When semantically related inputs are available (e.g., an image and its caption), we directly test Eq.~\ref{eq:rep-sim-enc} by using $h_t^\ell$, the LM's hidden state at position $t$ and layer $\ell$, as $M_{\text{LM}}(w_{1:t}^{z})$, and using cosine similarity for the similarity function.

We operationalize Eq.~\ref{eq:dom-readoutability-dec} via the \emph{logit lens}~\citep{logit-lens}, a simple training-free approach for interpreting the hidden states of a model. Transformer LMs produce the next-token distribution using $\operatorname{softmax}\left(Oh_t^{L}\right)$ (omitting the bias term) where $O$ is the output token embeddings (or ``unembeddings'') and $h_t^{L}$ is the final layer hidden state. Logit lens applies the same operation to the intermediate layers $\ell \in [L]$ to obtain $p^{\text{logitlens}}(\cdot \mid h_t^{\ell}):=\operatorname{softmax}\left(Oh_t^{\ell}\right)$. Logit lens has been found to produce  meaningful distributions that shed light on an LM's internal representations and computations.

Under the logit lens, $\operatorname{emb}(\cdot)$ in Eq.~\ref{eq:dom-readoutability-dec} is the output embedding of a token.
Using the dot product for $\operatorname{sim}(\cdot)$, Eq.~\ref{eq:dom-readoutability-dec} is equivalent to comparing the logit lens probabilities,
\begin{align} \label{eq:main-test}
p^{\text{logitlens}}\left(\tau^{z^\dom} \mid h^{\ell}_t\right) > p^{\text{logitlens}}\left(\tau^{z^\nodom} \mid h^{\ell}_t\right),
\end{align}
i.e., testing whether the probability of the continuation in the dominant language is more likely than the continuation in the original input data type.
Since the logit lens is tailored for probing out a single token, we usually consider short-enough verbalizations such that a single BPE token can reliability identify it, such as when the two verbalizations are two single words that are semantic equivalents. But we also consider longer future verbalizations when its first token unambiguously suggests one interpretation in that context, which allows more flexibility. Since
$\tau^{z^\nodom}$ is unavailable in many multimodal models without vocabulary tokens for $z^\nodom$; we only test Eq.~\ref{eq:rep-sim-enc} in those cases.

    \vspace{-2mm}
\section{Evidence of A Semantic Hub} \label{sec:identify}
    \vspace{-2mm}
We apply our tests across diverse data types and models: inputs in different languages, arithmetic expressions, code, formal semantic structures, visual inputs, and audio.
We consistently find evidence of a shared representation space in all cases.

\subsection{Multilingual} \label{sec:identify-multilingual}

Much past work has categorized language representations in multilingual LMs~\citepia{hua-etal-2024-mothello,alabi-etal-2024-hidden,tang-etal-2024-language,zhao2024largelanguagemodelshandle,zeng2024converginglinguafrancaevolution}.
\citet{wendler-etal-2024-llamas} recently find that when processing specific in-context learning (ICL) templates for highly synthetic lexical-level tasks (word repetition, word translation, etc.) in non-English languages, the intermediate hidden states of Llama-2 are closer to the unembeddings of English tokens than in the output language. This is consistent with our hypothesis, albeit constrained to a simple synthetic task and one LM.
We show that this shared representation space is a general property of LMs and also occurs  when processing naturalistic text.

\paragraph{Experiment 1: Representations are similar for translations.} Translation datasets enable a direct test for Eq.~\ref{eq:rep-sim-enc}, with semantically equivalent cross-lingual sentences as $w_{1:t}^{z_1}$ and $w_{1:t'}^{z_2}$ and a randomly chosen non-matching sentence as $u_{1:t''}^{z_2}$. We use the professionally-translated English-Chinese parallel sentences from \citet{gale2016} ($N=5260$). 
For each sentence pair, we use a template to transform each sentence 
and compute the representation cosine similarity for each layer, using the last token position as the sentence representation,
which has been shown to preserve sentential information~\citep{morris-etal-2023-text}.
We consider two English-dominant LMs, Llama-2 and Llama-3~\citep{dubey2024llama3herdmodels}, one Chinese-dominant LM, Baichuan-2~\citep{yang2023baichuan2openlargescale}, and one multilingual LM, BLOOM~\citep{workshop2023bloom176bparameteropenaccessmultilingual}, specifically the 7B/8B variants. See \S\ref{sec:appendix-multilingual} for more details.

\begin{figure}[t!]
    \begin{subfigure}[t]{0.328\textwidth}
        \centering
        \includegraphics[width=\textwidth]{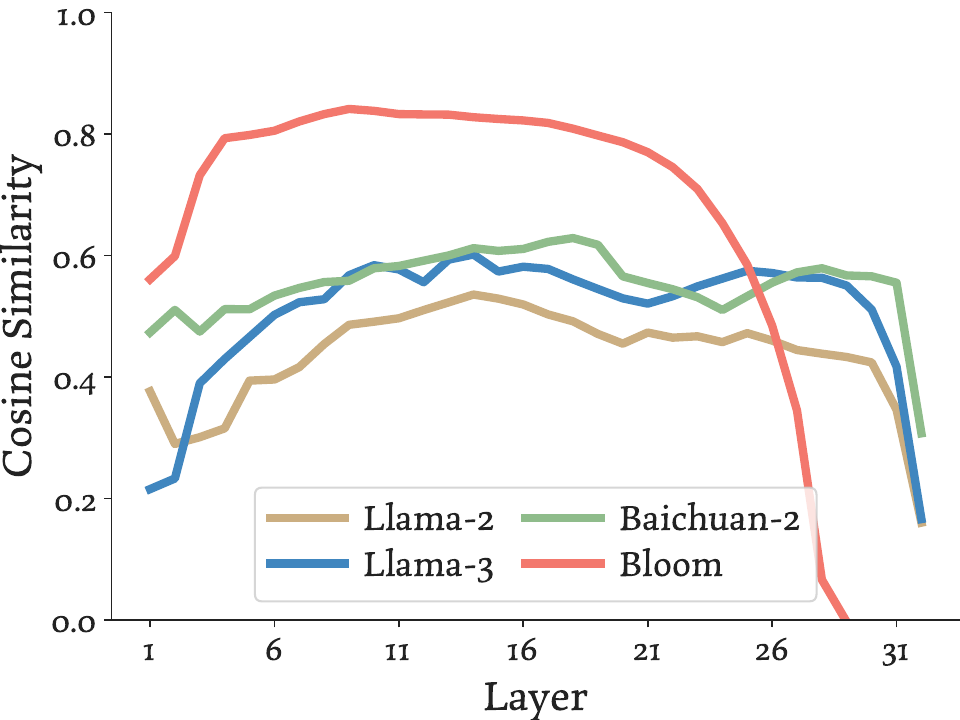}
        \caption{The cosine similarity of intermediate representations of English and Chinese parallel texts.}
        \label{fig:parallel-sentence}
    \end{subfigure}%
    ~~
    \begin{subfigure}[t]{0.328\textwidth}
        \centering
        \includegraphics[width=\textwidth]{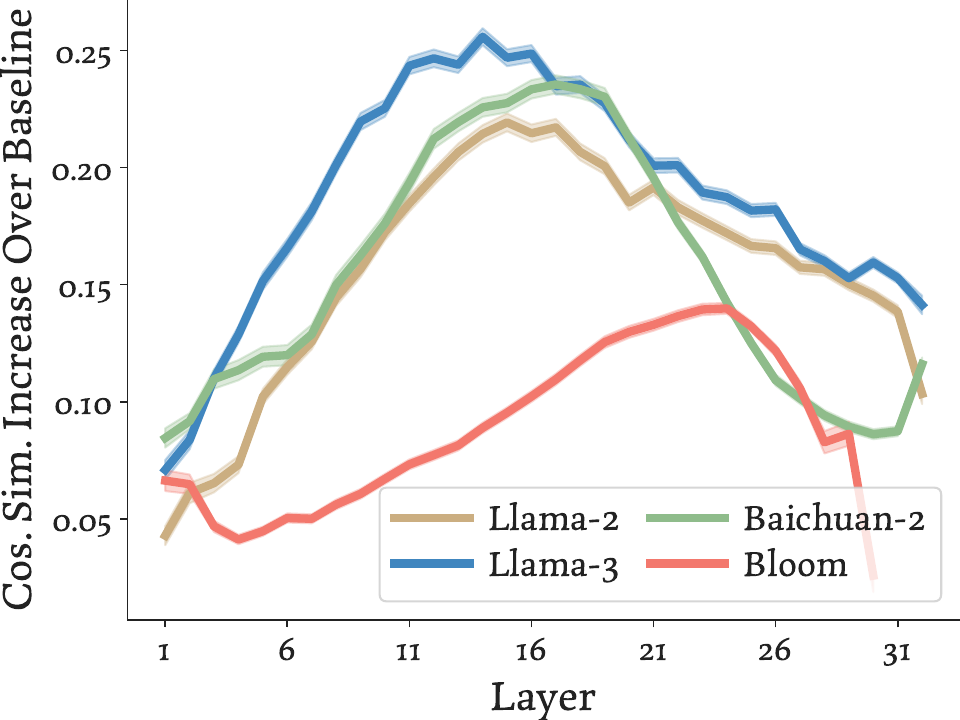}
        \caption{Same as (a), but subtracted by a baseline over non-parallel texts.}
        \label{fig:parallel-sentence-diff}
    \end{subfigure}%
    ~~
    \begin{subfigure}[t]{0.328\textwidth}
        \centering
        \includegraphics[width=\textwidth]{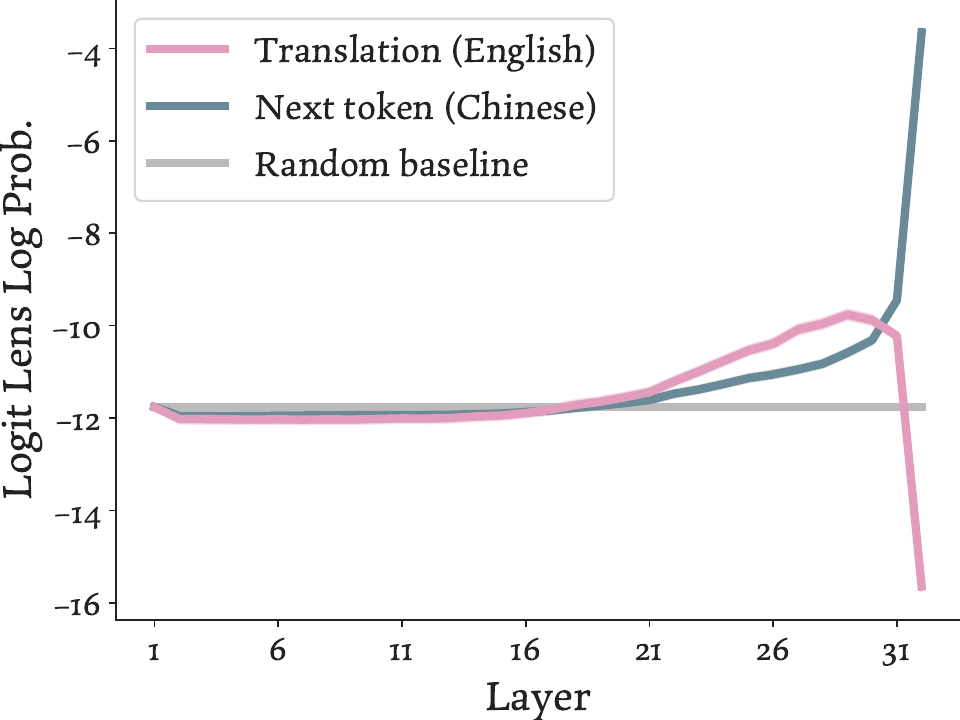}
        \caption{Llama-3 logit lens log prob. of parallel English vs. Chinese tokens when processing Chinese text.}
        \label{fig:llama3-translation-vs-actual}
    \end{subfigure}
    \caption{Results for the multilingual experiments. The 95\% CI is plotted in all. \textbf{Parallel texts have similar representations. Hidden states for Chinese texts are close to the unembedding of English tokens.}}
    \vspace{-3mm}
\end{figure}

Figure~\ref{fig:parallel-sentence} shows high raw cosine similarity. We also follow Eq.~\ref{eq:rep-sim-enc} and subtract the average similarity between non-matching sentence pairs as a baseline, separately for each layer. Figure~\ref{fig:parallel-sentence-diff} shows that the similarity between translations is higher than this baseline, most prominently in the middle layers.
These trends support the hypothesis that the middle layers act as the semantic hub of the LM.
Notably, this trend also exists for BLOOM which does not have a dominant pretraining language.

\begin{figure*}[t!]
    \vspace{-2mm}
\centering
\includegraphics[width=1\textwidth]{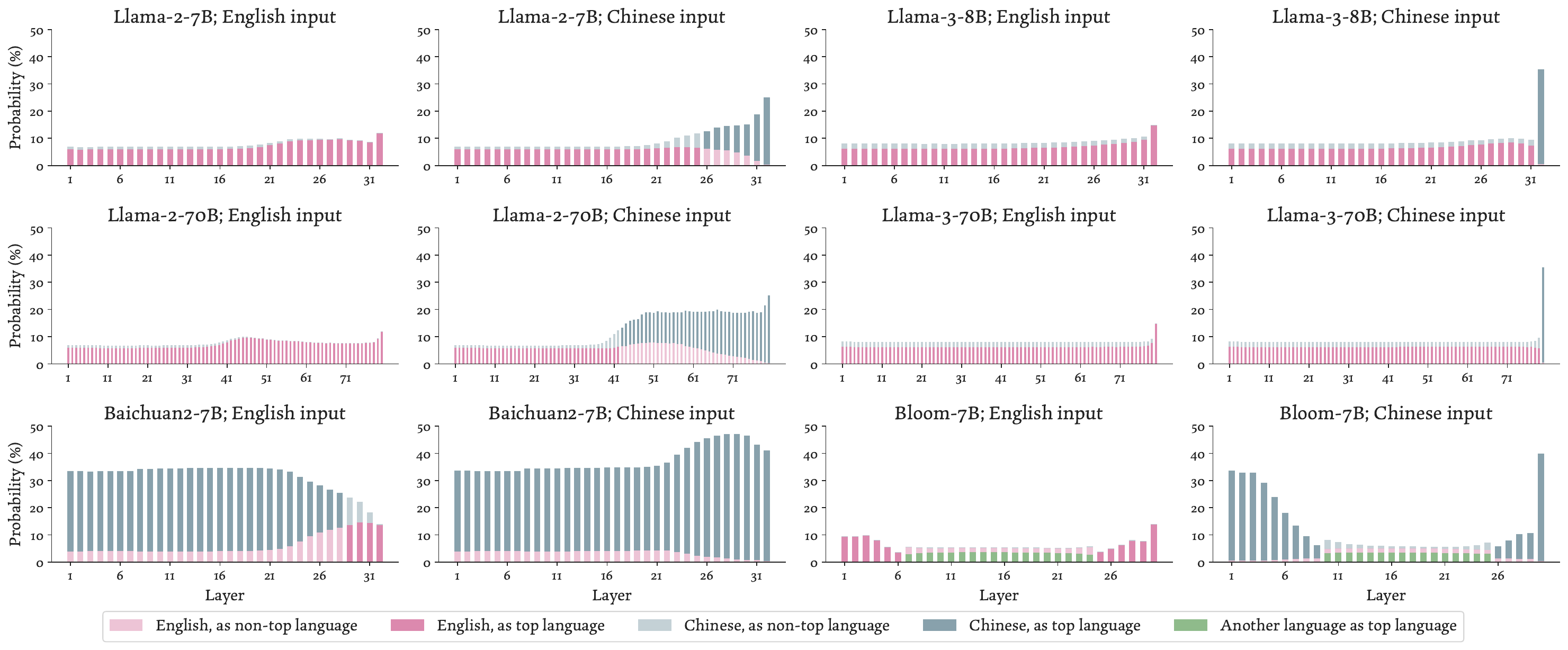}
    \caption{Language probabilities of English and Chinese (and the top language, when it is neither, which only happens for Bloom). \textbf{Regardless of the input language, the dominant LM language is more salient in the early-middle layers, and the input language is more salient in the final layers. Bloom does not have a clear intermediate latent language.}}
    \label{fig:percentage-output}
    \vspace{-3mm}
\end{figure*}

\paragraph{Experiment 2: Representations are anchored by semantically-equivalent dominant-language tokens.}
We next test Eq.~\ref{eq:main-test}, i.e., whether continuations in the dominant language have a higher probability than those in the input language at intermediate LM layers.
For the English-dominant Llama-3, we use 1,000 Chinese prefixes $w_{1:t}^{z^\nodom}$ from Wikipedia~\citep{wikidump} as input. For each, we take $\tau^{z^\nodom}$ to be the next Chinese token (i.e., $w_{t+1}^{z^{\nodom}}$) and 
$\tau^{z^\dom}$ to be the (first token of the) English translation of $w_{t+1}^{z^{\nodom}}$. See \S\ref{sec:appendix-multilingual} for more details.
Figure \ref{fig:llama3-translation-vs-actual} plots the logit lens probability for the two tokens as well as the uniform distribution probability.
In early layers, we cannot read out either token better than random chance. After layer 17, the model representations are substantially closer to the English token than the Chinese token until layer 31, showing that the model hidden space is indeed better scaffolded by English than Chinese. 

Next, we extend this analysis to consider global language-level trends $p^\ell(z)$ where $z$ is a language, at some layer $\ell$. We first compute $p(w\mid z)$, the token distribution under a language $z$, by running the LM tokenizer on the language-specific split of the mC4 dataset~\citep{xue-etal-2021-mt5}. We then use Bayes' rule to estimate $p(z\mid w)\propto p(w\mid z)p(z)$ with a uniform prior $p(z)$.\footnote{This prior obviously does not reflect the training language distribution, but in fact makes our trends even more salient, since using a real (or estimated) $p(z)$ would make $p(z\mid w)$ even larger for the dominant language.} We compute the probability of $h_t^\ell$'s belonging to each language as $p(z\mid h_t^\ell)\propto \sum_{w\in \mathcal{V}}p(z\mid w)p^{\text{logitlens}}(w\mid h_t^\ell)$. Finally, again with a  uniform prior assumption, we average $p(z\mid h_t^\ell)$ across tokens $t$ to obtain a language distribution $p^\ell(z)$. If our hypothesis that the shared representation space is better scaffolded by the dominant language is true, we expect the dominant language $z^{\dom}$ to have the highest probability across input languages in the middle layers.

Figure~\ref{fig:percentage-output} shows the probability of English and Chinese (and of the top language when it is neither) for each layer on 10,000 English/Chinese Wikipedia sentences.
When English-dominant models process Chinese text, \citeauthor{wendler-etal-2024-llamas}'s finding generalizes, where English dominates in the intermediate layers and Chinese only dominates in the final layers. On the Chinese-dominant LM, this trend flips: when processing English text, its intermediate layers are closer to the Chinese space and the final layers are closer to the English space.
For BLOOM, a multilingual model with a relatively balanced training language mixture, we do not see a clear dominating language in the intermediate layers; when we manually inspect the closest token, in most cases we observe symbols with no clear semantics (though this does not mean it does not have a unified representation space: see Exp. 1). 70B model trends are also highly similar to the 7B/8B ones.

\vspace{-2mm}
\subsection{Arithmetic} \vspace{-2mm}
\label{sec:identify-arithmetic}

\begin{figure}[t!]
    \begin{subfigure}[t]{0.328\textwidth}
        \centering
        \includegraphics[width=\textwidth]{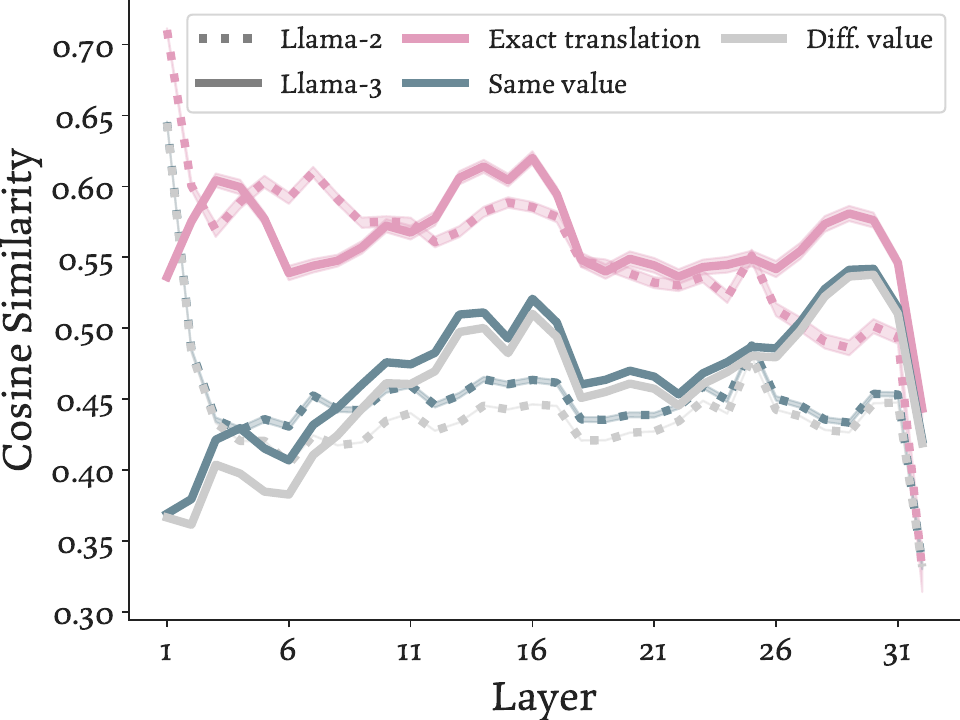}
        \caption{Cosine similarity between an arithmetic expression in Arabic numerals vs. English words, broken down into separate categories.}
        \label{fig:cosine_sim_arithmetic}
    \end{subfigure}%
    ~~
    \begin{subfigure}[t]{0.328\textwidth}
        \centering
        \includegraphics[width=\textwidth]{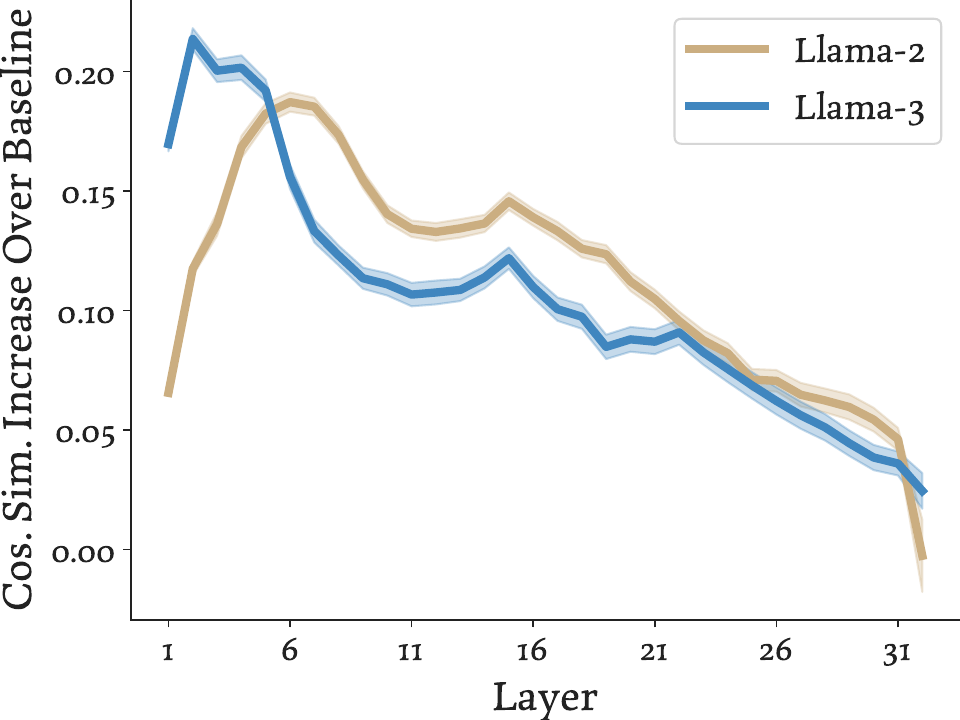}
        \caption{Same as (a), but only the exact translation similarities subtracted by the others.}
        \label{fig:cosine_sim_diff_arithmetic}
    \end{subfigure}%
    ~~
    \begin{subfigure}[t]{0.328\textwidth}
        \centering
        \includegraphics[width=\textwidth]{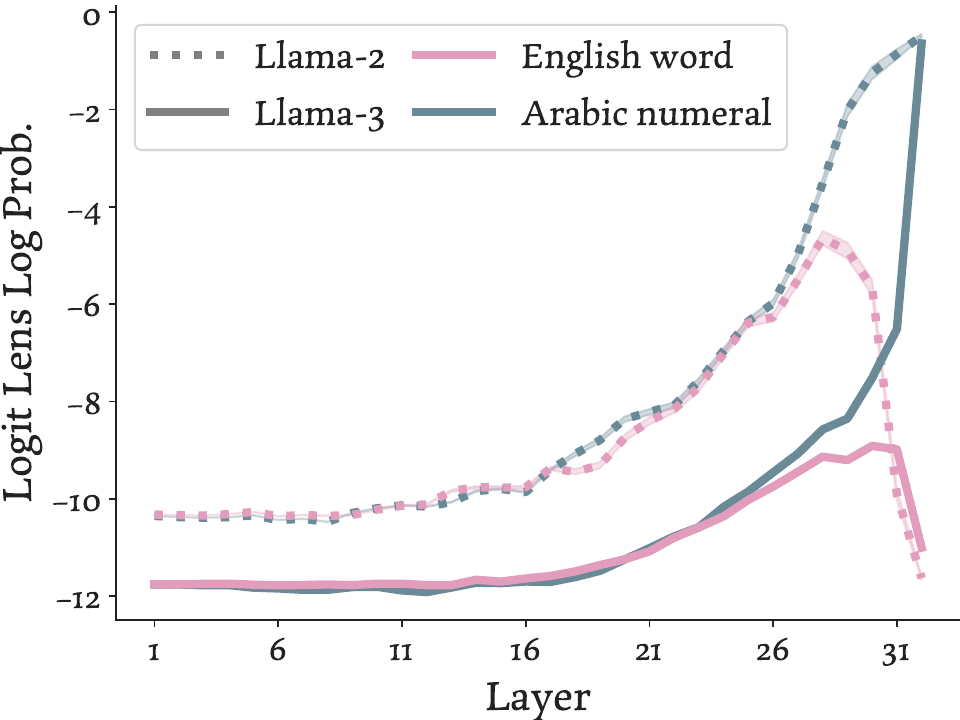}
        \caption{Logit lens log probability when predicting a number, between either the number itself or its English equivalent.}
        \label{fig:arithmetic}
    \end{subfigure}
    \caption{Results for the arithmetic experiments. The 95\% CI is plotted in all. \textbf{Expressions in Arabic numerals have similar representation as corresponding expressions in English, as well as the unembeddings of corresponding number words in English.}}
    \vspace{-3mm}
\end{figure}

We hypothesize that a similar trend exists when LMs process arithmetic expressions where they route to a shared space anchored by numerical words in English in intermediate layers. We consider simple expressions in the form of \token{a=b+c} or \token{a=b*c}; for simplicity, we restrict \token{a} and \token{b} to be at most two digits and \token{c} to be a single positive digit.

\paragraph{Experiment 1: Representations are similar for translations.}
Here, we only consider the right-hand side, \token{b+c} and \token{b*c}, as $w_{1:t}^{z_1}$ in Eq.~\ref{eq:rep-sim-enc}.
Like in the multilingual case, we translate them into English (e.g., \token{five plus three}) as $w_{1:t'}^{z_2}$, and evaluate the representation cosine similarity between every English expression and every numeric expression, throughout layers. We group the pairwise cosine similarities in three buckets: (1) exact translation (e.g., \token{5+3} and \token{five plus three}; $N=1123$), (2) non-exact but same value (e.g., \token{5+3} and \token{two plus six}; $N=13293$), and (3) different value ($N=1247836$). Figure~\ref{fig:cosine_sim_arithmetic} shows that exact translations have high cosine similarity, although this is to be expected since embeddings of numbers and their corresponding English words are near one another (thus even a bag-of-word-embeddings should also have high similarity). More interestingly, we that the similarities are still higher when the surface forms are distinct but the ``meaning'' of the expression (i.e., the value of the expression) is the same.
Next, like in \S\ref{sec:identify-multilingual}, we subtract the cosine similarities among non-translation pairs as a baseline ($u_{1:t'}^{z_2}$). Figure~\ref{fig:cosine_sim_diff_arithmetic} shows high similarity in the early-middle layers for translations over the baseline, but gradually decreasing to near 0.

\begin{wrapfigure}{R}{6.4cm}
    \vspace{-4mm}
    \begin{minipage}{0.47\textwidth}
    \centering
    \includegraphics[width=\textwidth]{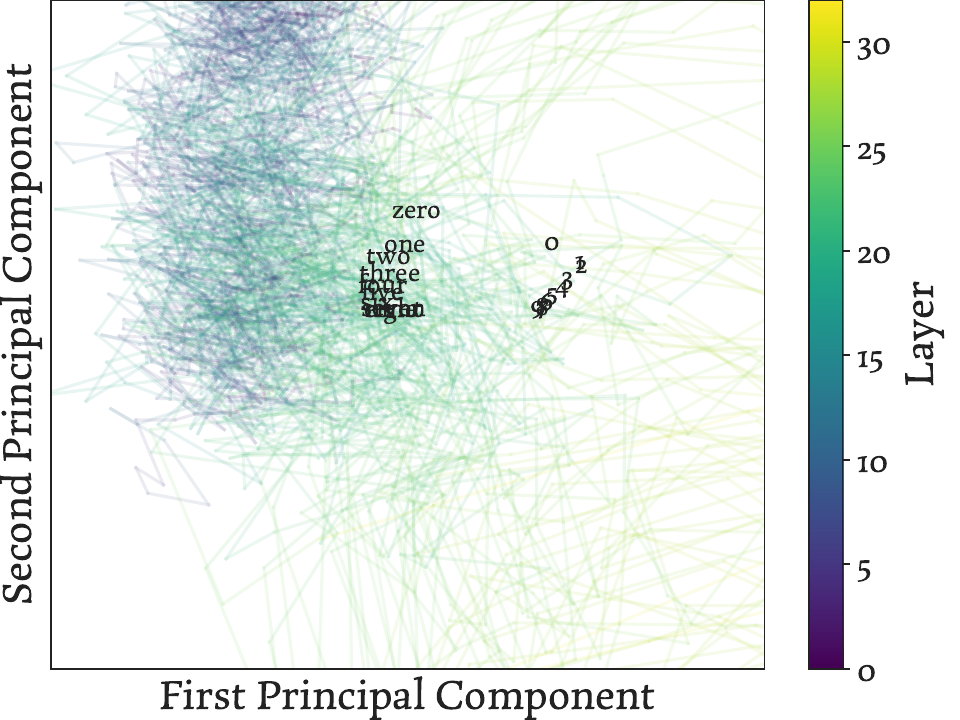}
    \caption{The Llama-3 hidden representation evolution when predicting a number, projected by PCA where the principal components are learned on the output embeddings of 20 number tokens, 10 in English and 10 numerals.}
    \label{fig:arithmetic-pca}
    \end{minipage}
    \vspace{-4mm}
\end{wrapfigure}

\paragraph{Experiment 2: Representations are anchored by semantically-equivalent English words.}
We hypothesize that, for some prefix such as \token{a=b+}, the intermediate representations $h_t^\ell$ are close to the English word for \token{c} that would make the equality hold (see Figure~\ref{fig:overview}).
First, we randomly sample 100 such prefixes and take the representation of the last token at all layers. For each prefix, we plot the representation evolution throughout layers using PCA, as well as the unembeddings of numbers in English $\tau^{z^\dom}$ vs. numerals $\tau^{z^\nodom}$. Figure~\ref{fig:arithmetic-pca} shows that the representations indeed go through the space occupied by the English words in intermediate layers.
Next, we repeat our logit lens experiments, inspecting the log probability of the following numeral token vs. its English version ($N=1123$). Figure~\ref{fig:arithmetic} shows that the two tokens have similar log probability until around layer 25, after which the numeral token dominates.

\vspace{-2mm}
\subsection{Code} \vspace{-2mm}
\label{sec:identify-code}
\begin{wrapfigure}{R}{8.3cm}
\vspace{-6mm}
    \begin{minipage}{0.6\textwidth}
    \includegraphics[width=\textwidth]{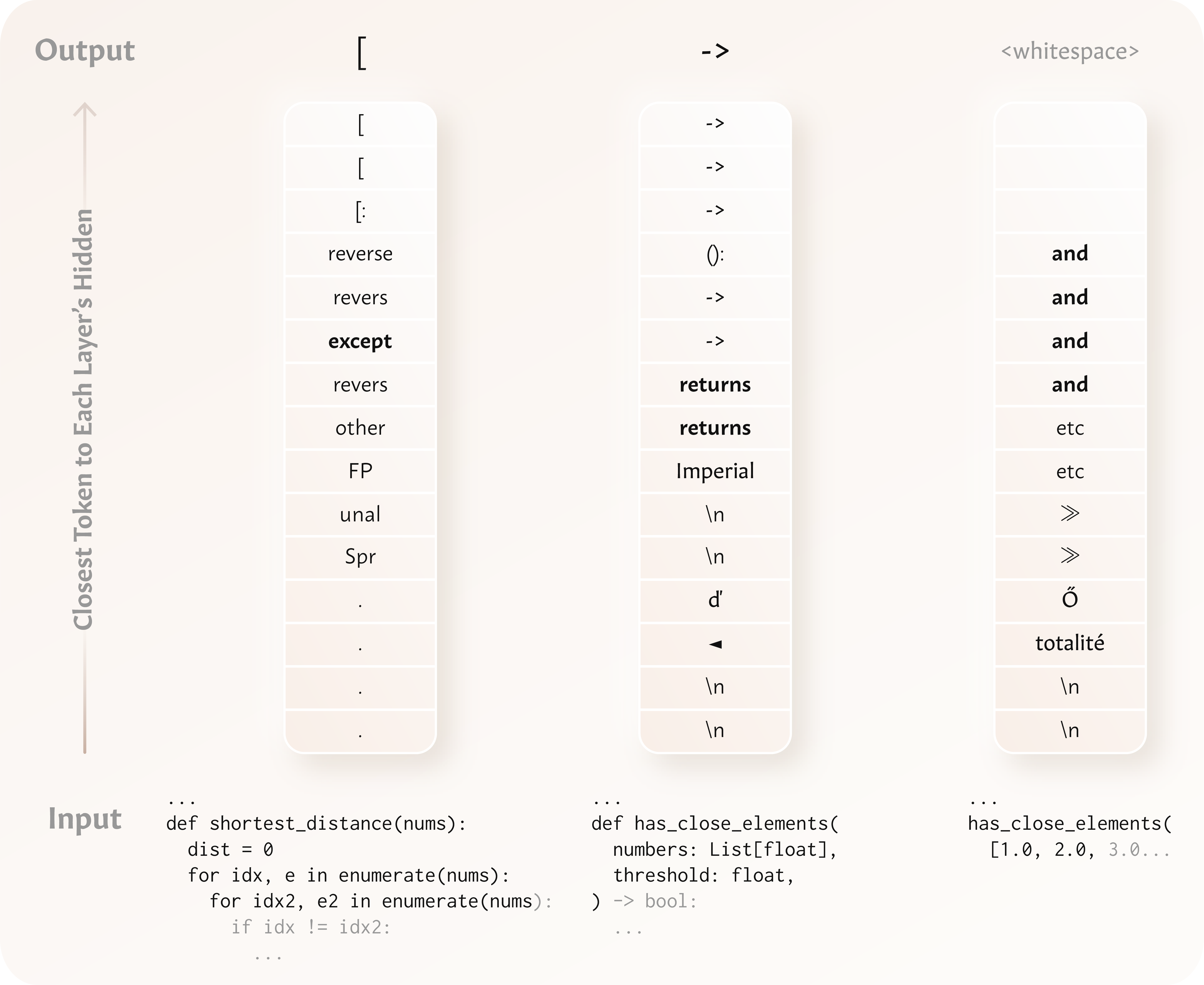}
    \caption{Logit lens analysis on Llama-2 processing Python programs. For every other layer, we show the closest token (sometimes whitespace) to the hidden states \emph{before} the grayed-out texts. \textbf{The model tends to verbalize the future prediction in English that corresponds to the code continuations (in gray)}.}
    \label{fig:code-examples}
    \vspace{-4mm}
    \end{minipage}
\end{wrapfigure}
Many recent LMs are trained on code corpora~\citepia{dubey2024llama3herdmodels, geminiteam2024geminifamilyhighlycapable}. We find that they similarly process code input by projecting it into a unified representation space shared with regular language tokens. Figure~\ref{fig:code-examples} shows some examples, where the LM in the intermediate layers tends to verbalize the future in free-form English, unconstrained by program syntax. E.g., in the first program, given the Python prefix \token{... for idx, e in enumerate(numbers): for idx2, e2 in enumerate(numbers}, instead of the ground-truth continuation in Python \token{): if idx != idx2:}, the most salient intermediate token is \token{except}, likely attempting to predict in English \token{\textcolor{gray}{(for each element in numbers)} except if it is equal to idx}. Similarly, in a list literal expression \token{[1.0, 2.0,} instead of continuing in Python \token{ 3.0}, it predicts \token{and}, which is a natural way to continue in English. 
In these cases, it is difficult to obtain semantically equivalent English-Python prefix pairs like in \S\ref{sec:identify-multilingual} for testing Eq.~\ref{eq:rep-sim-enc}\footnote{We may argue that functions and their specifications constitute such pairs, but we found that their correspondence is often too abstract and non-exact to manifest as similar representations.}, so we only test Eq.~\ref{eq:main-test} across a few targeted cases in Python below.

\begin{figure}
\vspace{-4mm}
     \begin{subfigure}[t]{0.35\textwidth}
        \centering
        \includegraphics[width=\textwidth]{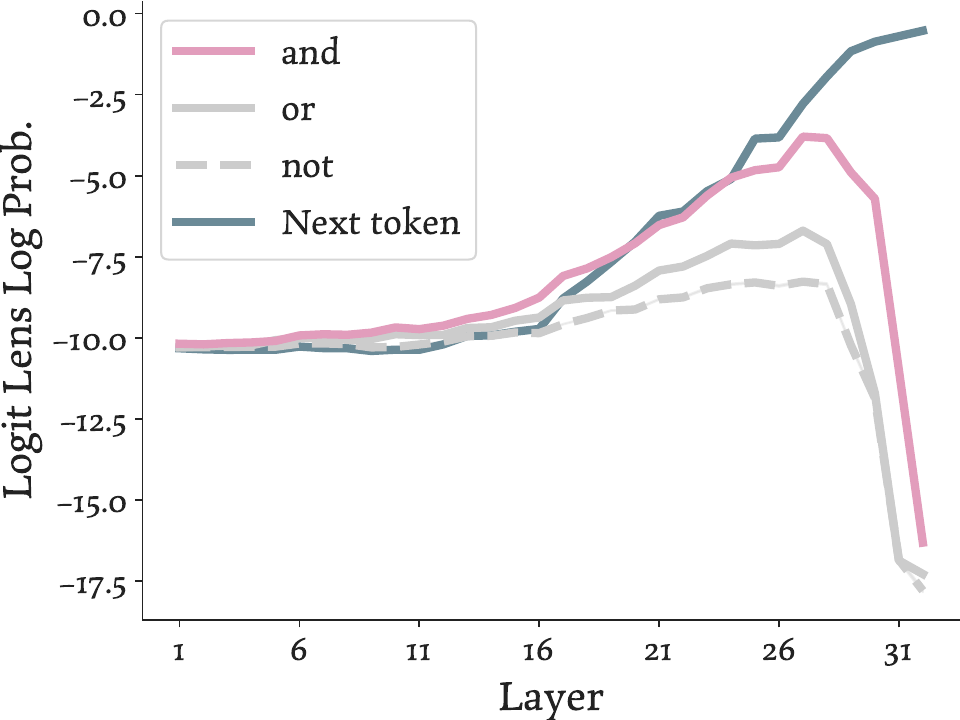}
        \caption{Llama-2 logit lens log probabilities (and the 95\% CI) at commas in Python list literals, of the English \token{and} token (and baseline tokens) vs. the actual next token in the program, from MBPP.}
        \label{fig:code-if}
    \end{subfigure}%
    ~~
    \begin{subfigure}[t]{0.63\textwidth}
        \centering
        \includegraphics[trim=0cm 4cm 0cm 3.5cm, clip, width=\textwidth]{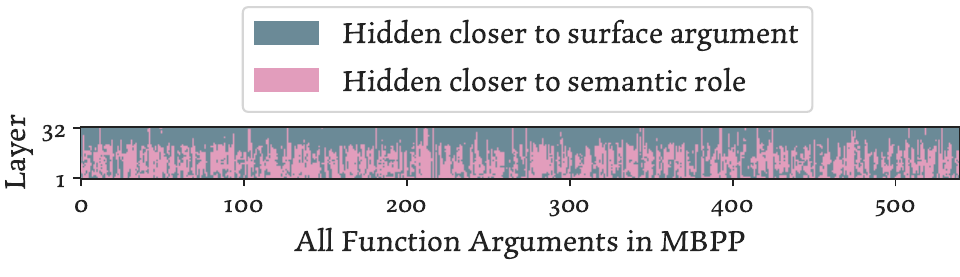}
        \caption{The distance between Llama-2 hidden states when predicting a function argument, to the unembedding of the argument's name (its semantic role) vs. the actual argument expression, in MBPP.}
        \label{fig:code-semantic-role}
    \end{subfigure}
    \caption{Results for the code experiments. \textbf{Code expressions are close to semantically meaningful free-form English words in early-middle layers, such as \token{and} in list literals and the argument's semantic role in function calls; in the final layers, the representation converges to the context-constrained Python token.}}
    \vspace{-3mm}
\end{figure}

\vspace{-2mm}
\paragraph{Experiment 1: Representations are anchored by semantically-equivalent English words: list literals.}
We systematically test the list case, where $h_t^l$ is the hidden state after processing \token{,} during list processing. We use $\tau^{z^\dom}=$\token{and} and further take $\tau^{z^\nodom}$ to be the actual next token.
Figure~\ref{fig:code-if} shows that this trend holds  on all such commas in the MBPP dataset~\citep{austin2021program} ($N=6923$, including unit tests): as expected, in the final layers, the representation is closer to the ground truth next token's unembedding, and closer to \token{and} in the middle layers. We also show the probability with two other tokens, \token{or} and \token{not}, as baselines, both of which are lower than \token{and}.

\vspace{-1mm}
\paragraph{Experiment 2: Representations are anchored by semantically-equivalent English words: function call arguments.}
Function arguments have names in the definition, such as \token{range(start, end, step)}; but when invoked, they are filled with actual context-appropriate expressions. We call the argument names ``semantic roles'', and the context-specific expressions the ``surface forms'', inspired by thematic relations in linguistics~\citepia{fillmore,jackendoff1974semantic}. Like in the second example in Figure~\ref{fig:overview}, we show that LMs predict the arguments by first ``thinking'' about their semantic role ($\tau^{z^\dom}$) and then instantiating with surface-constrained expressions ($\tau^{z^\nodom}$).
We extract all function calls and arguments from MBPP with simple filtering, resulting in 540 arguments (see  \S\ref{sec:appendix-identify-code} for details). For each argument, we use the logit lens to inspect the hidden states $h_t^\ell$ at the preceding token (\token{(} or \token{,}). For each argument, Figure~\ref{fig:code-semantic-role} visualizes if the semantic role or the surface form is closer to each layer's hidden state. The semantic role ($\tau^{z^\dom}$) dominates for the early to middle layers,
and only in the final layers do the representations converge towards the surface form argument ($\tau^{z^\nodom}$).

\vspace{-2mm}
\subsection{Formal Semantics} \vspace{-2mm}
\label{sec:identify-semantics}

\begin{figure}[t!]
    \begin{subfigure}[t]{0.328\textwidth}
        \centering
        \includegraphics[width=\linewidth]{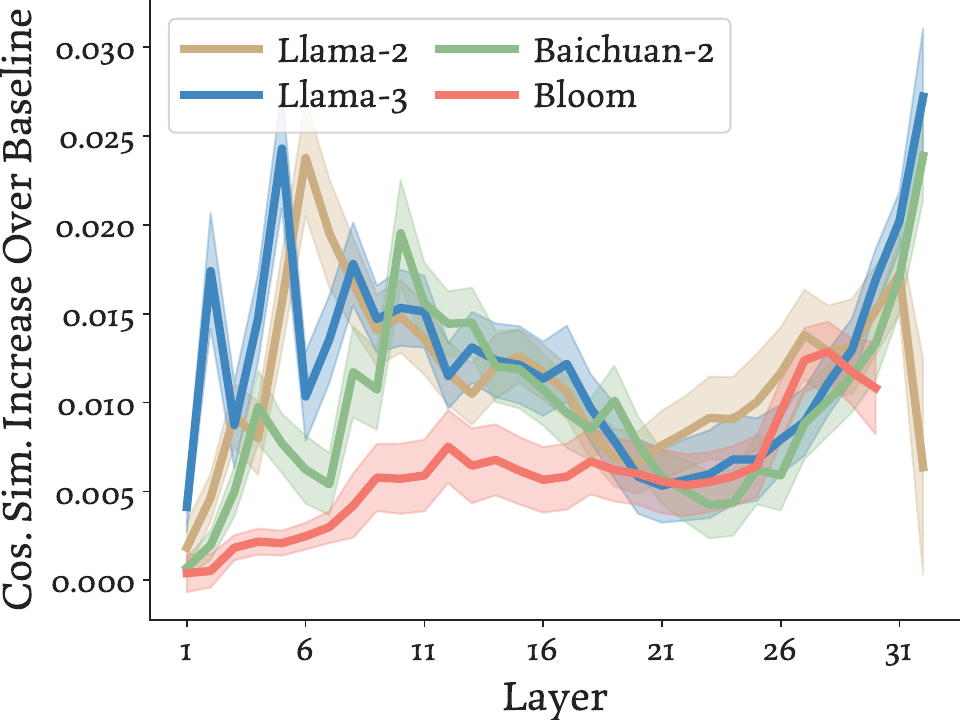}
        \caption{Similarity between a sentence and its semantic structure, subtracted by a baseline over non-parallel texts.}
        \label{fig:cogs-unfiltered}
    \end{subfigure}%
    ~~
    \begin{subfigure}[t]{0.328\textwidth}
        \centering
        \includegraphics[width=\linewidth]{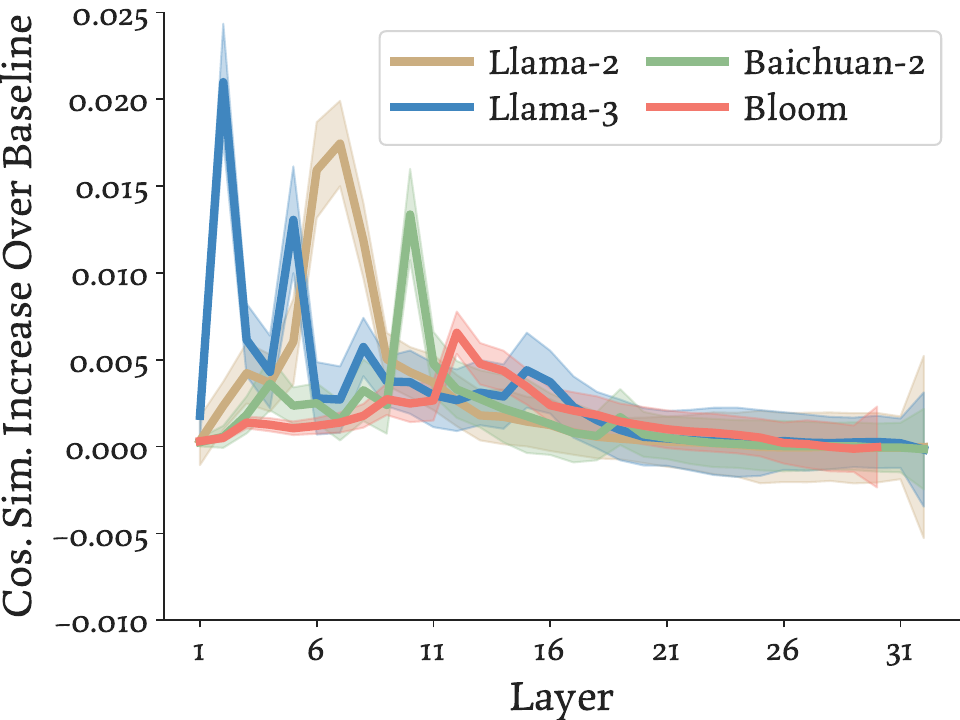}
        \caption{Similar as (a), but the baseline is computed by swapping the positions of names in the semantic structure.}
        \label{fig:cogs-filtered}
    \end{subfigure}%
    ~~
    \begin{subfigure}[t]{0.328\textwidth}
        \centering
        \includegraphics[width=\textwidth]{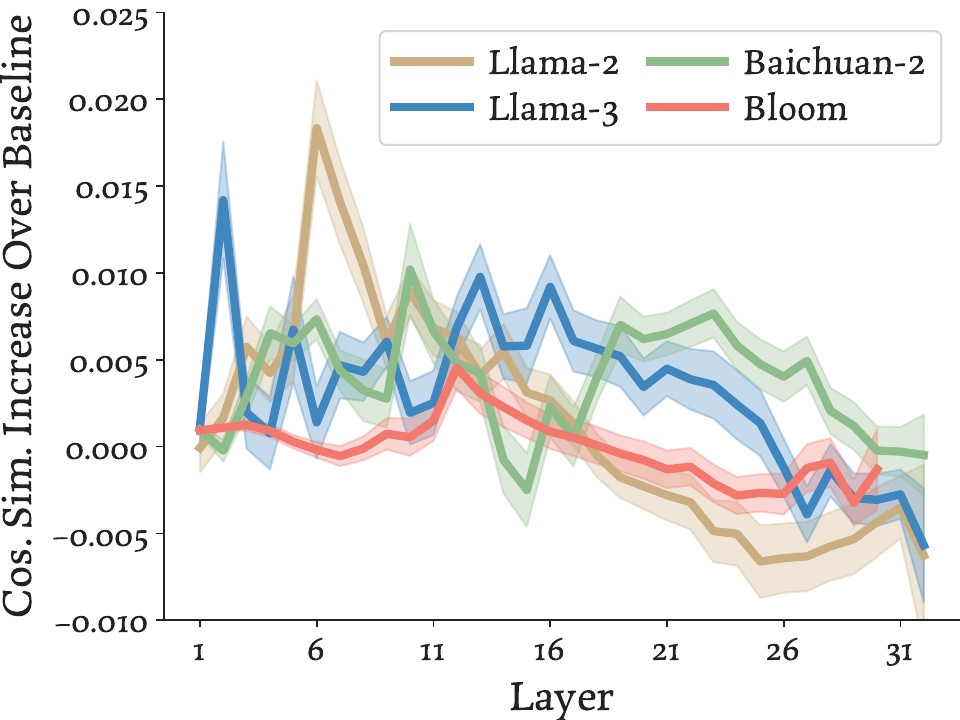}
        \caption{Same as (b), but we shuffle the predicates in the semantic structure to ensure robustness.}
        \label{fig:cogs-shuffled}
    \end{subfigure}
    \caption{Representation similarity experiments for formal semantics. \textbf{A sentence and its semantic structure have high representation similarity}, even with strict controls.}
\end{figure}

\begin{figure}[t!]
    \begin{subfigure}[t]{0.32\textwidth}
        \centering
        \includegraphics[width=\textwidth]{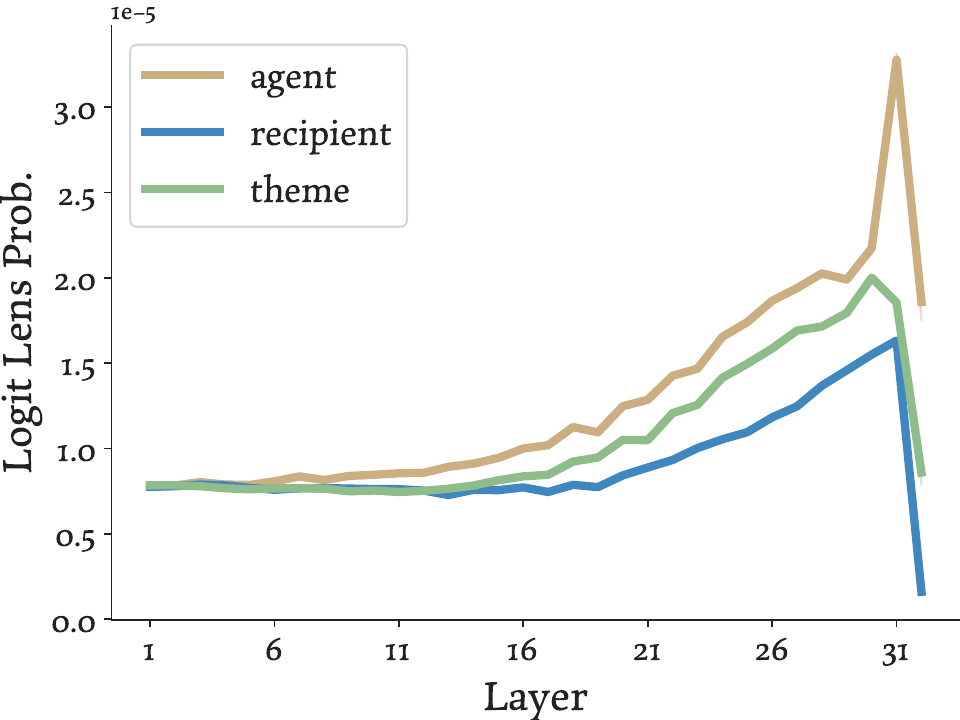}
        \caption{Agent}
        \label{fig:cogs_semantic_role_goldagent_prob}
    \end{subfigure}%
    ~
    \begin{subfigure}[t]{0.32\textwidth}
        \centering
        \includegraphics[width=\textwidth]{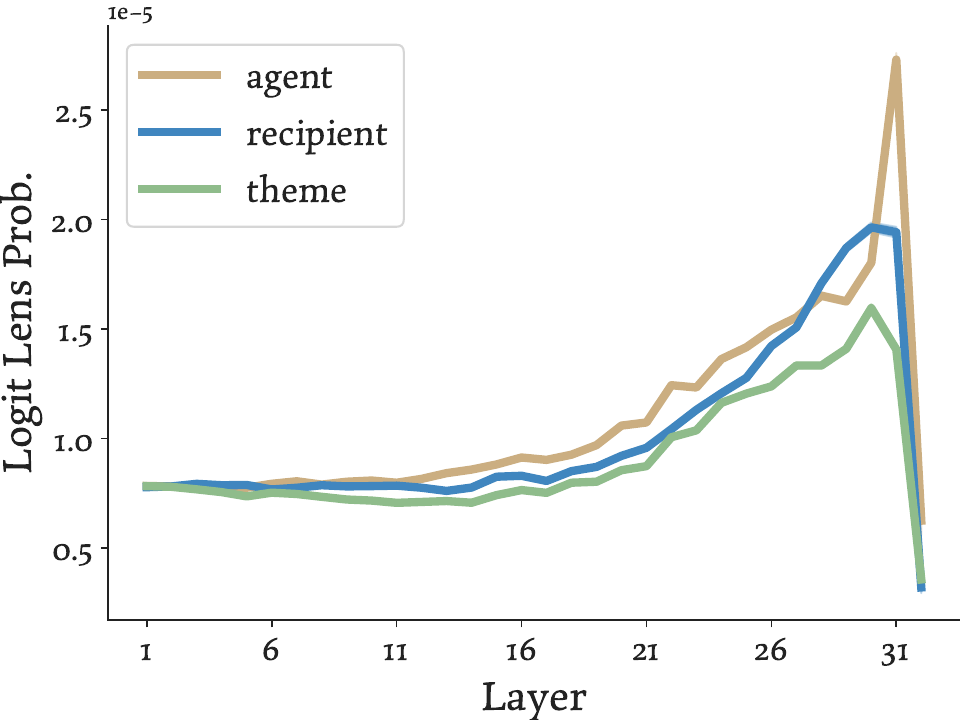}
        \caption{Recipient}
        \label{fig:cogs_semantic_role_goldrecipient_prob}
    \end{subfigure}%
    ~
    \begin{subfigure}[t]{0.32\textwidth}
        \centering
        \includegraphics[width=\textwidth]{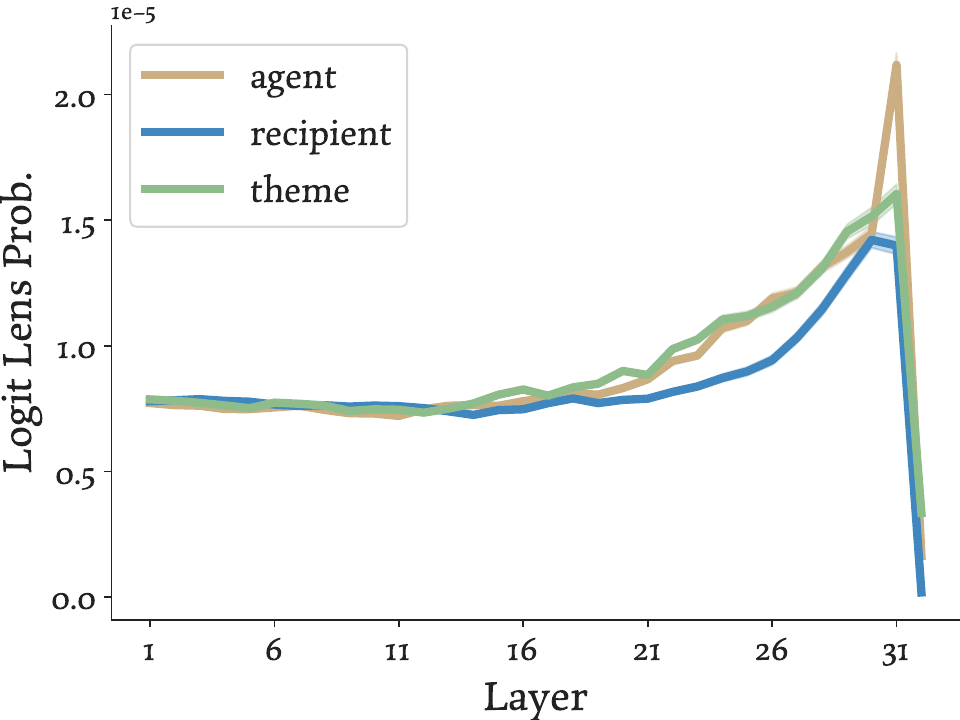}
        \caption{Theme}
        \label{fig:cogs_semantic_role_goldtheme_prob}
    \end{subfigure}%
    \\
    \begin{subfigure}[t]{0.32\textwidth}
        \centering
        \includegraphics[width=\textwidth]{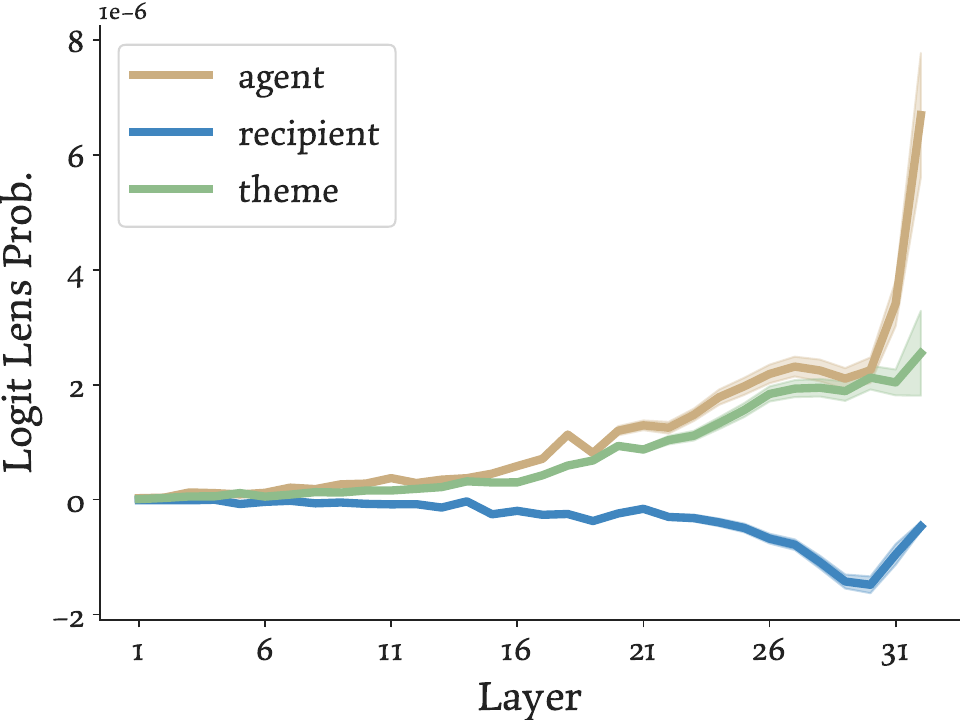}
        \caption{Agent}
        \label{fig:cogs_semantic_role_goldagent_prob_adj}
    \end{subfigure}%
    ~
    \begin{subfigure}[t]{0.32\textwidth}
        \centering
        \includegraphics[width=\textwidth]{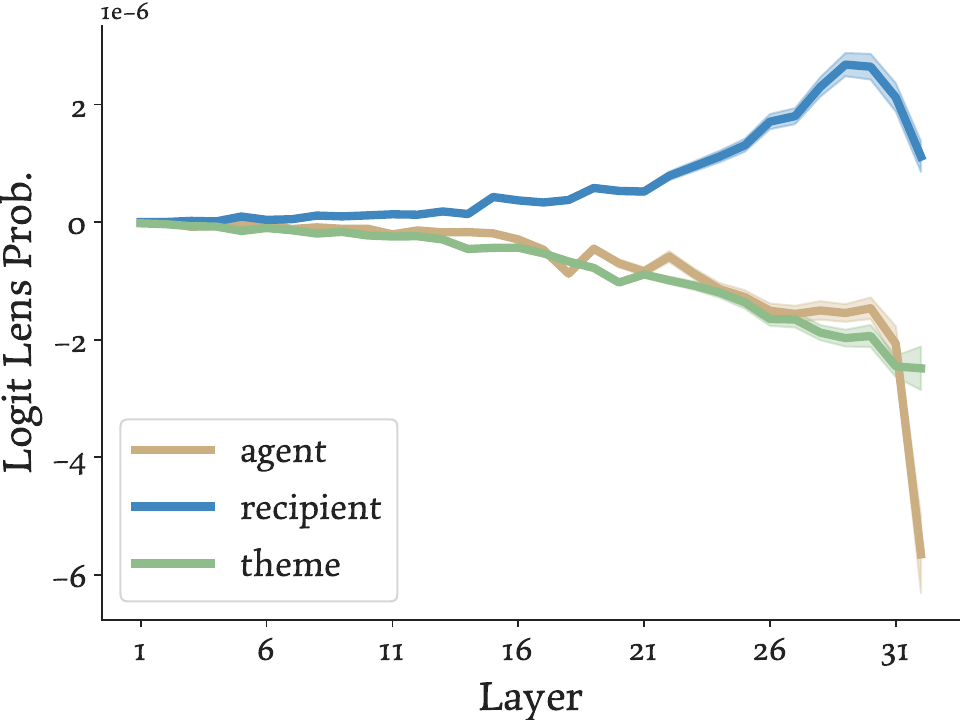}
        \caption{Recipient}
        \label{fig:cogs_semantic_role_goldrecipient_prob_adj}
    \end{subfigure}%
    ~
    \begin{subfigure}[t]{0.32\textwidth}
        \centering
        \includegraphics[width=\textwidth]{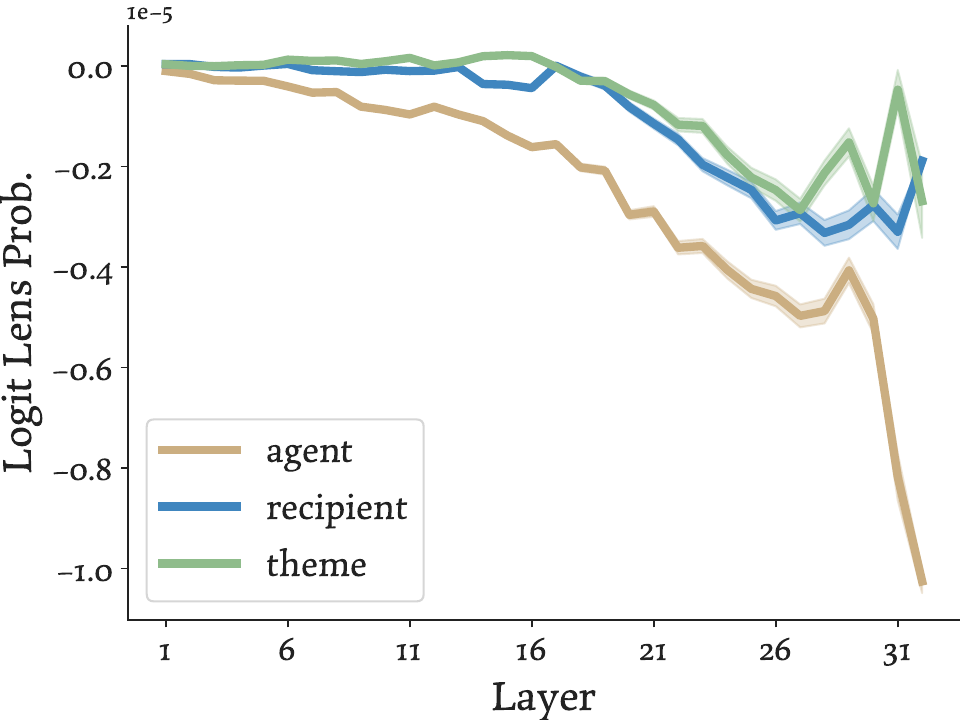}
        \caption{Theme}
        \label{fig:cogs_semantic_role_goldtheme_prob_adj}
    \end{subfigure}
    \caption{\textbf{Top}: Logit lens probability of the thematic roles of verb arguments, grouped by the ground-truth argument thematic role in each subplot, for Llama-3. \textbf{Bottom}: The same quantity, subtracted by the average probability as the baseline, separately for each role. The 95\% CI is plotted in all. \textbf{The prefix representation before an argument with a particular thematic role has high similarity with the unembedding of that role in English}, when the baseline is adjusted for.}
    \vspace{-3mm}
\end{figure}

Much probing work has shown that semantic information can be probed out from LM's hidden states~\citepia{tenney-etal-2019-bert,sift,li-etal-2021-implicit}. We show that this manifests without a learned probe. We use the COGS data~\citep{kim-linzen-2020-cogs}, which contains synthetically generated English sentences and their semantic structures (in a fairly standard format rooted in the Neo-Davidsonian tradition~\citep{Parsons1990-PAREIT}).\footnote{It also has a similar number of active vs. passive sentences, thus providing a clean testbed since one cannot simply use word order to predict the thematic role.} E.g., the sentence \token{Eleanor sold Evelyn the cake.} has the representation \token{*cake(x4); sell.agent(x1, Eleanor) AND sell.recipient(x1, Evelyn) AND sell.theme(x1, x4)}.

\paragraph{Experiment 1: Representations are similar between a sentence and its semantic structure.}
Like in \S\ref{sec:identify-multilingual} and \S\ref{sec:identify-arithmetic}, Figure~\ref{fig:cogs-unfiltered} shows that the representation of a sentence is closer to its semantic structure than a non-matching baseline. As in the arithmetic expression case however, there is a confounder: this could be due to surface lexical overlap between the two (e.g., \token{Eleanor}, \token{Evelyn}, \token{cake}, and \token{sell}/\token{sold} in our example) rather than a deep understanding of their equivalence. To control for this, we find COGS sentences with two proper names like the above example ($N=2233$), swap their positions in the semantic structure such that it no longer corresponds to the sentence, and yet the lexical overlap is unchanged.
With this stronger baseline (where bag-of-words models would do no better than chance), we still see in Figure~\ref{fig:cogs-filtered} that semantically matching pairs have more similar representations.
On top of this, we further control potential positional confounders of predicates in the semantic structure by randomly swapping their positions, and results in Figure~\ref{fig:cogs-shuffled} have the same trend.
We also note that BLOOM, arguably the ``weakest'' model (smaller pretraining set and older architecture) does not do much better than chance, potentially suggesting that this representation strategy correlates with model capacity.

\paragraph{Experiment 2: Word representations are correlated with their thematic roles in semantics theory.}
Formal semantic structures are naturally not a dominant data type, and we hence do not expect Eq.~\ref{eq:main-test} to hold.
Instead, we perform a logit-lens-style analysis that still tests Eq.~\ref{eq:rep-sim-enc}, but with finer granularity, focusing on \emph{token}-level representation similarity, specifically the arguments of verbs and their thematic roles. For example, \token{Eleanor} in the earlier example has the role \token{agent} and \token{Evelyn} is the \token{recipient}. We expect that, when predicting an argument, the hidden states are more similar to the corresponding thematic role than non-matching ones. We again only consider proper names in COGS, excluding those that start a sentence, as in this case there is no context in which to predict the semantic role. This results in 3257 agents, 2583 recipients, and 714 themes.
To avoid any memorization effects arising from the LMs' being potentially trained on COGS, we further randomly replace the proper names with another one in COGS and make sure the new sentence does not appear in the dataset.

For each proper name with a given thematic role, we look at the logit lens probability of all roles. Figures~\ref{fig:cogs_semantic_role_goldagent_prob} to \ref{fig:cogs_semantic_role_goldtheme_prob} show that, for Llama-3, the role probabilities tend to peak in the middle-late layers and drop down in the final layers, a familiar trend. Nevertheless, the corresponding role token does not always receive the highest probability. We believe this is because Llama-3 has a ``prior'' that is closer to the word \token{agent}, which almost always has the highest probability. So we adjust for this prior by subtracting from each curve a baseline probability, separately for each role and for each layer, that is the average role probability across all instances. Figures~\ref{fig:cogs_semantic_role_goldagent_prob_adj} to \ref{fig:cogs_semantic_role_goldtheme_prob_adj} show this ``posterior'': when predicting the argument with a given thematic role, the corresponding thematic role token is always the closest to the intermediate representations.
We do not observe these trends in Llama-2.

\vspace{-2mm}
\subsection{Visual Input} \label{sec:identify-visionlanguage}

\begin{figure}
     \begin{subfigure}[t]{0.328\textwidth}
        \centering
        \includegraphics[width=\textwidth]{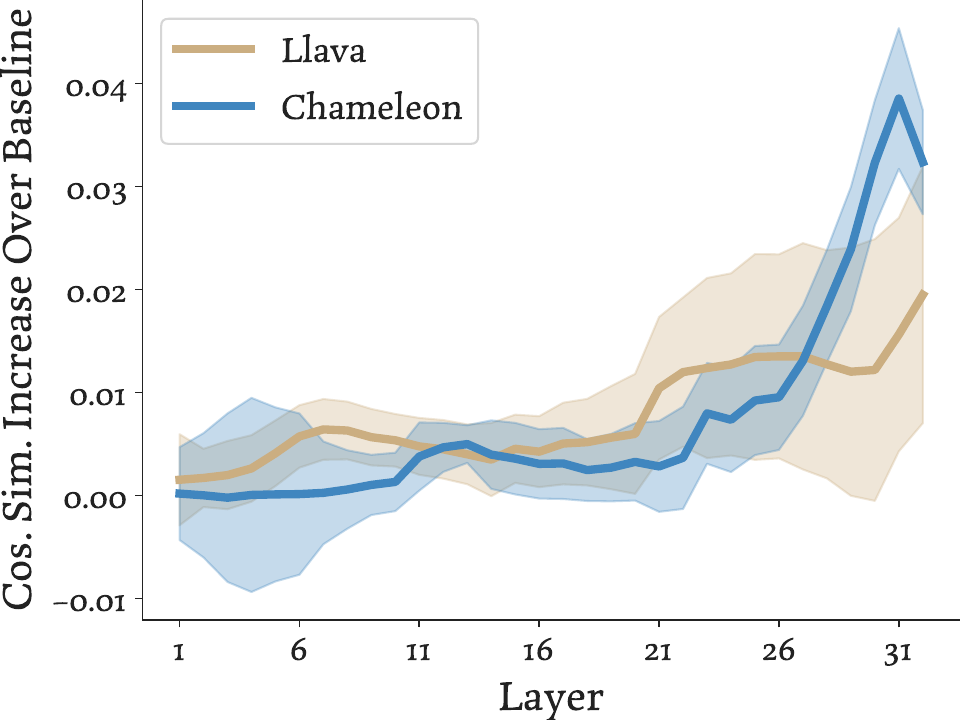}
        \caption{The cosine similarity difference between intermediate representations of matching images and captions, over non-matching ones.}
        \label{fig:caption-representation-similarity}
    \end{subfigure}%
    ~~
    \begin{subfigure}[t]{0.328\textwidth}
        \centering
        \includegraphics[width=\textwidth]{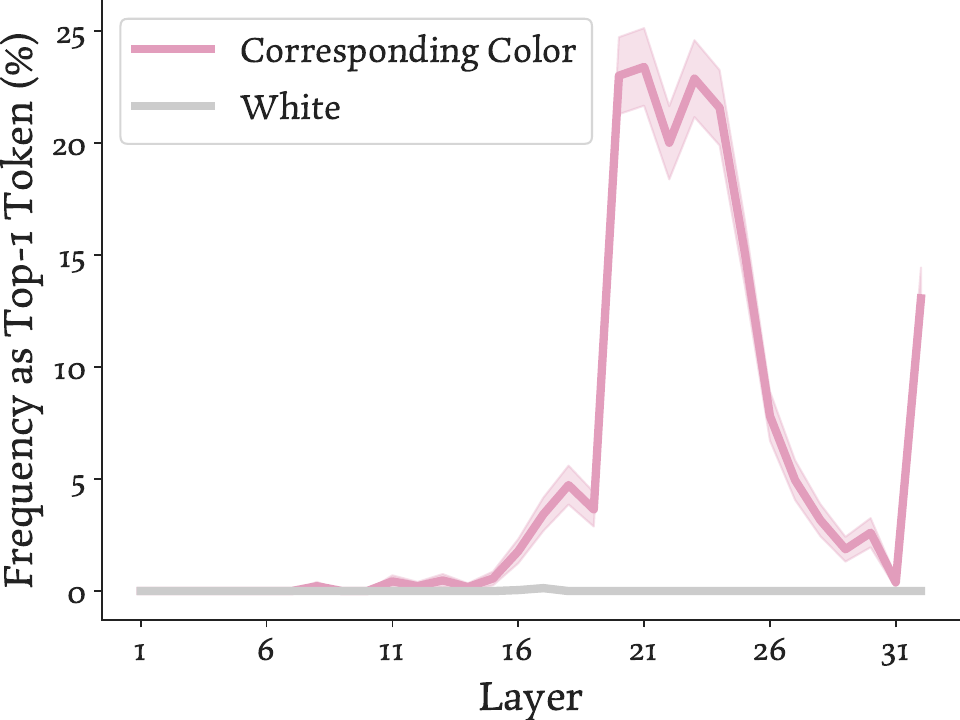}
        \caption{The frequency of the closest token to LLaVA's hidden states describing the image color, against a baseline using \token{white}.}
        \label{fig:color-unembedding}
    \end{subfigure}%
    ~~
    \begin{subfigure}[t]{0.328\textwidth}
        \centering
        \includegraphics[width=\textwidth]{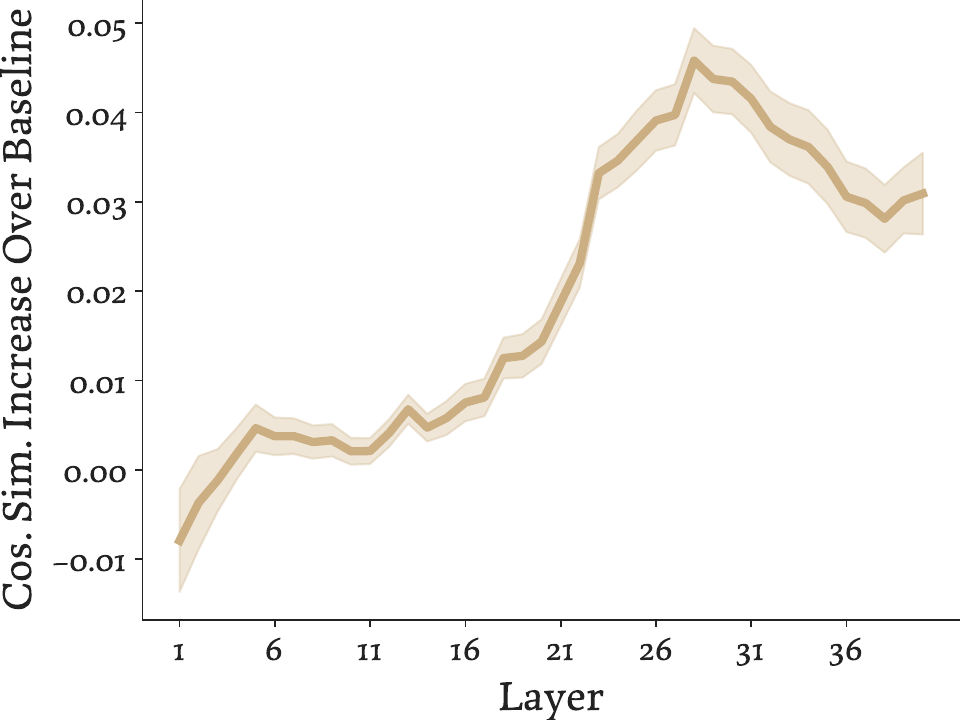}
        \caption{The cosine similarity difference between intermediate SALMONN representations of matching audios and labels, over non-matching ones.}
        \label{fig:audio-representation-similarity}
    \end{subfigure}
    \caption{Results for the multimodal experiments. The 95\% CI is plotted in all. \textbf{Model representation of (a) visual and (c) audio inputs and their textual labels are similar. Furthermore, in (b), model representations of individual color patch tokens are close to English words for those colors.}
    More results are in \S\ref{sec:appendix-logitlens}.}
    \vspace{-3mm}
\end{figure}

Past work has investigated the representation of \emph{separately trained} vision and text models, often finding that their representation spaces are similarly structured and alignable~\citepia{merullo2022linearly,li2023implications,huh2024platonic}. 
We show that when trained together, vision-language models learn to project both modalities into a joint representation space.
Current vision-language models typically represent images by segmenting them into patches, embedding them into ``image tokens'', and then feeding them into the transformer model along with other text tokens~\citepia{uio,uio2,liu2023llava}. We hypothesize that the intermediate representations of the image patches are close to the corresponding language tokens that describe the scene.
Experimental details are in \S\ref{sec:appendix-identify-vision-language}.

\paragraph{Experiment 1: Representations are similar between an image and its caption.} 
Though not constituting exact semantic equivalence,
an image paired with its caption provides one possible test for Eq.~\ref{eq:rep-sim-enc}.
We take 1000 images and corresponding captions in the MSCOCO dataset~\citep{coco} and measure their hidden states cosine similarity in LLaVA-7B~\citep{liu2023llava} and Chameleon-7B~\citep{Chameleon_Team_Chameleon_Mixed-Modal_Early-Fusion_2024}.
Again, we subtract the average similarity between non-matching image-caption pairs as a baseline, separately for each layer. Figure~\ref{fig:caption-representation-similarity} shows that semantically matching inputs across modalities are closer to one another than would be expected from chance, as in the translation experiments.
Importantly, these models do not explicitly optimize cross-modal representation similarity (unlike in CLIP~\citep{clip}), and such similarity only emerges through autoregressive training.

As mentioned in \S\ref{sec:method-new}, we do not perform a test of dominant data type anchoring (Eq.~\ref{eq:main-test}) due to a lack of non-text tokens in the vocabulary. But, like in \S\ref{sec:identify-semantics}, we perform a fine-grained image-patch-level analysis using the logit lens.
As a toy setting for illustration, we inspect LLaVA's representations of pure color images, specifically those in red, green, blue, and black. Figure~\ref{fig:color-unembedding} shows that, in up to more than 20\% of the time in the intermediate layers (averaged across the patches and the four colors, $N=2304$), the closest token is the corresponding color word (out of all vocabulary tokens).
\S\ref{sec:logit-lens-multimodal} details more comprehensive experiments where we similarly find an alignment between the patches and the caption, as well as between a patch and its semantic segmentation label.

\vspace{-2mm}
\subsection{Audio} \label{sec:identify-audio}
\vspace{-2mm}

Audio is another modality that is often modeled jointly with text~\citepia{uio2,jasu,gong2024listen}, and we perform similar experiments using SALMONN~\citep{tang2024salmonn}, an audio-text model.
We use the VGGSound dataset~\citep{Chen20VGGS} which contains 10-second audio clips with labels, e.g., \token{duck quacking} or \token{playing cello}.

\paragraph{Experiment 1: Representations are similar between audio and its label.}
We study the representation cosine similarity between an audio and its label description, and subtract from it a baseline which is the average similarity between non-matching pairs, separately for each layer. 
On 1000 samples from VGGSound, we see in Figure~\ref{fig:audio-representation-similarity} that semantically matching audios and labels have more similar representations in the intermediate layers.

Like for visual inputs, we do not investigate the dominant data type anchoring effect, but instead use the logit lens to confirm that the representation alignment also occurs on a token level, in \S\ref{sec:logit-lens-audio}.
\vspace{-2mm}
\section{Intervening in the Semantic Hub} 
\vspace{-2mm}\label{sec:intervention}

Prior work has argued that interpretability results should be tested under a causal framework, to ensure that the observation is not a vestigial byproduct of model training that has no actual effect on model behavior~\citepia{vig2020investigating,ravichander-etal-2021-probing,elazar2021amnesic,chan2022causal}. In this section, we show that the semantic hub does causally affect model output.
Specifically, semantically transforming hidden representations according to (dominant) English representations leads to predictable behavior changes in non-dominant data types.
For different experiments, we use different kinds of intervention, and we explain our design decisions at the end of this section.
We report hyperparameters and further experimental details in \S\ref{sec:appendix-intervention}.

\begin{table*}[t!]
    \centering
    \caption{\label{tab:actadd}
    Steering Llama-3's output sentiments using trigger words in English vs. the input language (either Spanish or Chinese). We report the mean sentiment, disfluency (perplexity), and relevance of the continuation, as well as the standard deviation across 10 seeds.  \textbf{Cross-lingual steering is consistently successful, sometimes even more than monolingual steering, without substantial damage in text fluency and relevance.}
    }
    \footnotesize
    \begin{tabular}{ccc|ccc}
        \toprule
        Text Lang. & Steering Dir. & Steering Lang. & Sentiment & Disfluency ($\downarrow$) & Relevance ($\uparrow$) \\
        \midrule
        \multirow{5}[3]{*}{Spanish} & None & None & 0.143$_{\pm0.022}$ & \phantom{0}7.35$_{\pm1.19}$ & 0.861$_{\pm0.002}$ \\
        \cmidrule{2-6}
        & \multirow{2}{*}{$\downarrow$} & Spanish & 0.125$_{\pm0.034}$ & 10.54$_{\pm2.39}$ & 0.842$_{\pm0.004}$ \\
        && English & 0.139$_{\pm0.026}$ & \phantom{0}8.75$_{\pm2.20}$ & 0.857$_{\pm0.002}$ \\
        \cmidrule{2-6}
        & \multirow{2}{*}{$\uparrow$} & Spanish & 0.175$_{\pm0.035}$ & \phantom{0}7.98$_{\pm2.04}$ & 0.856$_{\pm0.002}$ \\
        && English & 0.159$_{\pm0.026}$ & \phantom{0}7.35$_{\pm1.01}$ & 0.859$_{\pm0.003}$ \\
        \cmidrule{1-6}
         \multirow{5}[3]{*}{Chinese} & None & None & 0.178$_{\pm0.030}$ & 11.06$_{\pm3.12}$ & 0.869$_{\pm0.004}$ \\
        \cmidrule{2-6}
        & \multirow{2}{*}{$\downarrow$} & Chinese & 0.152$_{\pm0.040}$ & 10.78$_{\pm2.66}$ & 0.866$_{\pm0.005}$ \\
        && English & 0.161$_{\pm0.029}$ & 11.36$_{\pm1.13}$ & 0.864$_{\pm0.004}$ \\
        \cmidrule{2-6}
        & \multirow{2}{*}{$\uparrow$} & Chinese & 0.153$_{\pm0.034}$ & 11.12$_{\pm3.12}$ & 0.870$_{\pm0.004}$ \\
        && English & 0.179$_{\pm0.032}$ & 10.90$_{\pm3.25}$ & 0.869$_{\pm0.003}$ \\
        \bottomrule        
    \end{tabular}
    \vspace{-5pt}
\end{table*}
\paragraph{Multilingual.} 
Past work has shown that (monolingual) interventions in the middle layers can steer the output of LMs in predictable ways~\citepia{subramani-etal-2022-extracting,turner2024activationadditionsteeringlanguage,rimsky-etal-2024-steering}. If the English-dominant LMs have a shared representation space, we should be able to intervene on this space in English even when processing other languages~\citep{dumas2024how}. We use a popular hidden space intervention technique, Activation Addition (ActAdd; \citealp{turner2024activationadditionsteeringlanguage}), which works by: (1) taking a pair of contrasting steering words that semantically represent the steering effect (e.g., ``\texttt{Good}'' and ``\texttt{Bad}'' for sentiment steering), (2) taking their hidden state difference at an intermediate layer, and 
(3) scaling and adding the steering vector to the hidden states for the original forward pass of the regular generation process, at the same layer, at the beginning of the sequence.
We generalize their sentiment-steering experiment cross-lingually. See \citet{turner2024activationadditionsteeringlanguage} for details.

We consider two non-dominant languages, Spanish and Chinese, and take 1000 prefixes each from the InterTASS dataset (Spanish; \citealp{DazGaliano2018TheDO}) and the multilingual Amazon reviews corpus (Chinese; \citealp{keung-etal-2020-multilingual}), and generate continuations either without modifications or intervened using ActAdd.
As the steering vector, we use the difference between positive vs. negative sentiment trigger words, in the appropriate direction. Specifically, we use \token{Good} and \token{Bad} for English, \token{Bueno} and \token{Malo} for Spanish, and ``\chin{好}'' and ``\chin{坏}'' for Chinese.
In addition to sentiment evaluation, we also measure the generation fluency and compute the relevance of the generation with the prefix using trained models, following \citet{turner2024activationadditionsteeringlanguage}.
Ideally, the intervention should achieve the desired sentiment without hurting text fluency and relevance, and the  English intervention should be just as effective as the text language intervention (see \S\ref{sec:appendix-intervention-multilingual} for more details).

Because we take intermediate layer representations of the steering words (step (1)), if the semantic hub is language-agnostic, we expect similar cross-lingual representations and in turn similar steering effects across steering languages.
Table~\ref{tab:actadd} shows that this is indeed true for Llama-3:
ActAdd in the text language is often effective, achieving the intended effect on sentiment, with usually only a statistically insignificant decrease in fluency and relevance. This aligns with the English-only findings in \citet{turner2024activationadditionsteeringlanguage}.
And intervening in English is similarly effective as using the text language.
Table~\ref{tab:actadd-llama2} (appendix) shows the results for Llama-2, with very similar trends.

\begin{figure}[t!]
    \begin{subfigure}[t]{0.245\textwidth}
        \centering
        \includegraphics[width=\textwidth]{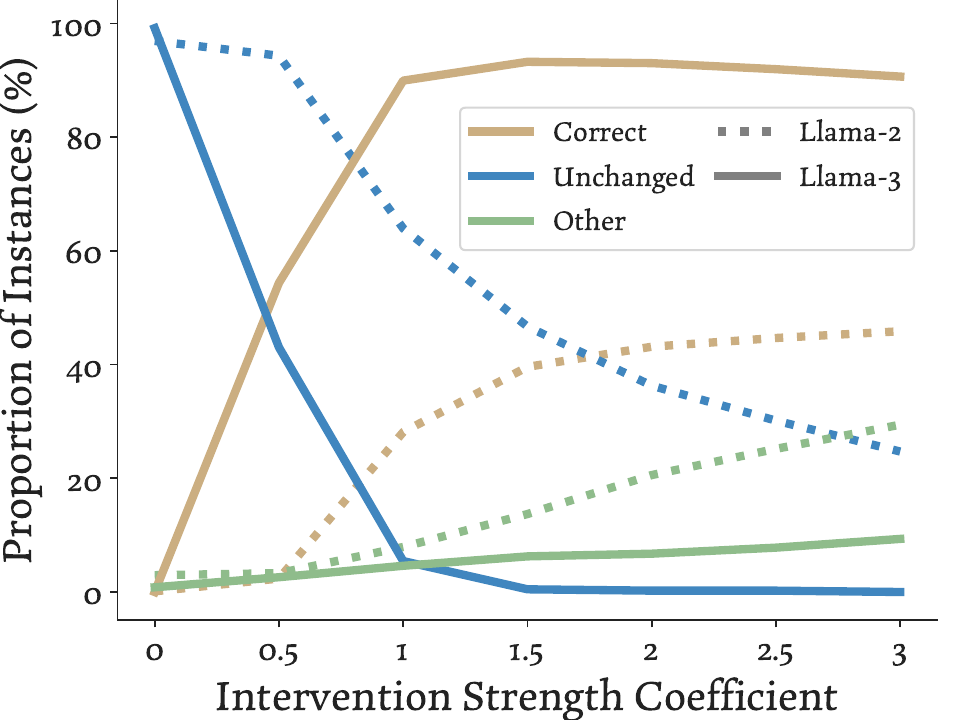}
        \caption{Steering arithmetic expressions' results to a different value.}
        \label{fig:arithmetic_intervention}
    \end{subfigure}%
    ~
    \begin{subfigure}[t]{0.245\textwidth}
        \centering
        \includegraphics[width=\textwidth]{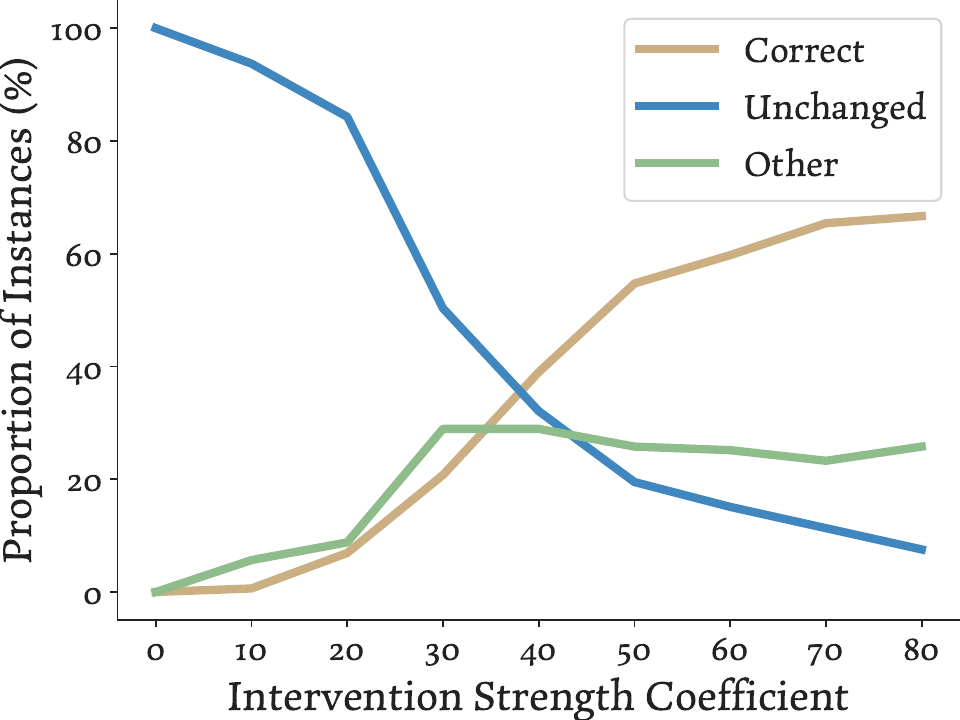}
        \caption{Steering a single-argument \token{range(end)} call to be predicted as double argument.}
        \label{fig:code-intervention}
    \end{subfigure}%
    ~
    \begin{subfigure}[t]{0.245\textwidth}
        \centering
        \includegraphics[width=\textwidth]{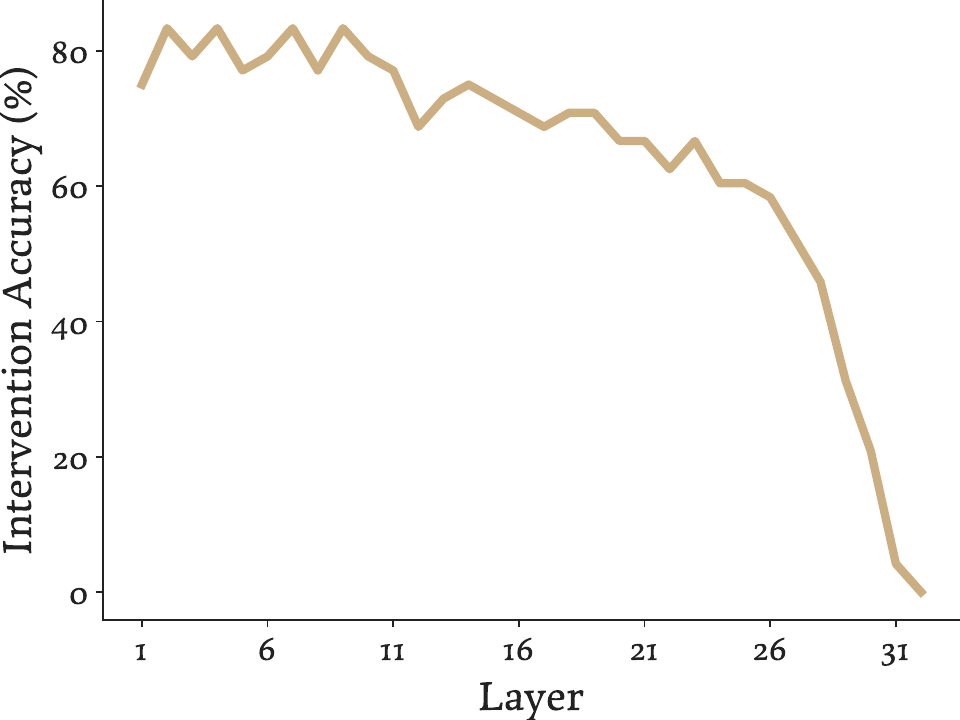}
        \caption{Replacing image representations of a color with English tokens of another color, and expecting the model to predict the latter.}
        \label{fig:color-intervention}
    \end{subfigure}%
    ~
    \begin{subfigure}[t]{0.245\textwidth}
        \centering
        \includegraphics[width=\textwidth]{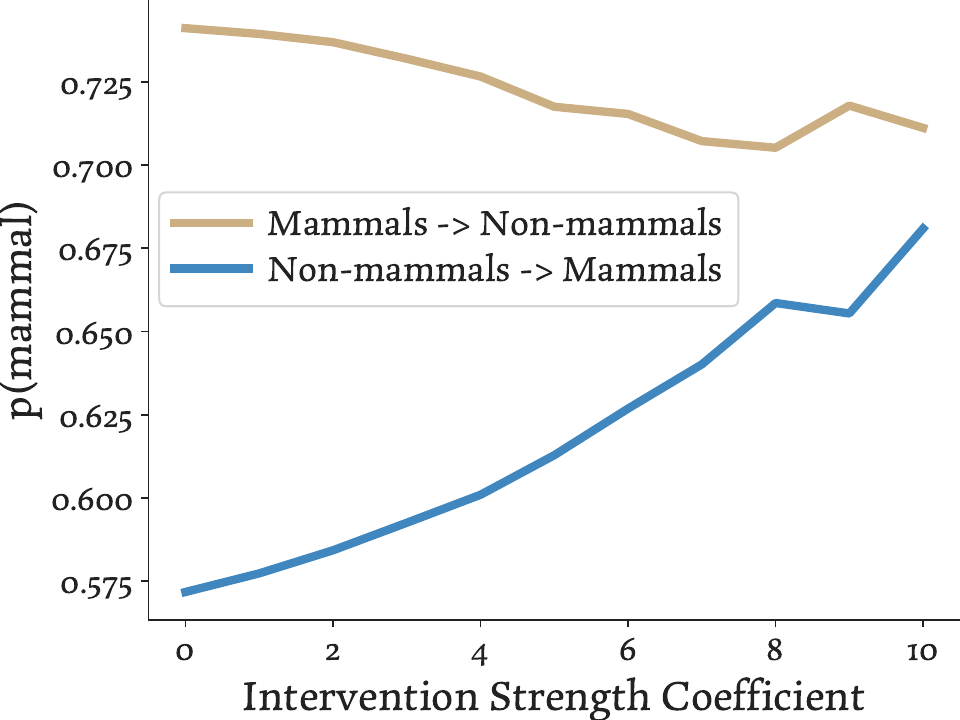}
        \caption{Steering mammal sounds to be predicted as non-mammal sounds using English words of non-mammals; vice versa.}
        \label{fig:audio-intervention}
    \end{subfigure}
    \caption{For (a) arithmetic (Llama-2 and -3), (b) code (Llama-2), (c) images (Chameleon), and (d) audio (SALMONN), steering model output using English words, for various intervention strengths ((a), (b), (d)) and layers ((c)). (a)-(c) measure successfulness with the proportion of instances steered to the correct output, and (d) measures the probability of predicting mammals. \textbf{Overall, intervening in the unified representation space in English reliably leads to desired model output changes.}}
    \vspace{-3mm}
\end{figure}

\paragraph{Arithmetic.}
We perform intervention using our arithmetic expressions in \S\ref{sec:identify-arithmetic}, for example \token{4=1+3}.
We intervene by attempting to modify the token after \token{+} to be one smaller, e.g. \token{2} here, and expect this to not only lead the model to output \token{2} instead of \token{3}, but also fundamentally affects the model's reasoning process and causes the model to patch this error with an additional suffix \token{+1}, i.e., \token{4=1+2+1}.
We use ActAdd except for adding the intervention vector (e.g., \token{three} -- \token{two}) only at the position of \token{+}.\footnote{Another difference is that we do not use the hidden representation after seeing e.g. \token{three}, because that usually represents the \emph{next} token. Instead, we use a prefix that uniquely determines the number, e.g. \token{Eight equals to five plus}, and take the last token hidden representation, which \emph{is} supposed to represent \token{three}.}
For all addition expressions in \S\ref{sec:identify-arithmetic} ($N=846$), we perform such intervention at an intermediate layer (25 for Llama-3 and 30 for Llama-2) and measure how often this leads to the model correctly outputting the decremented number followed by \token{+1}, versus unchanged, or changed to some other output. Figure~\ref{fig:arithmetic_intervention} shows that, as the intervention coefficient (i.e., the scaling constant of the vector) increases, this procedure leads to the expected output for up to $>90\%$ of the instances.

 \label{sec:intervention-code}
\paragraph{Code.} Based on our ``semantic role'' observation in \S\ref{sec:identify-code}, we intervene using the \token{range} function. %
We focus on two overloaded versions of \token{range}: \token{range(start, end)} and \token{range(end)}. If the semantic hub causally affects model output, then we can intervene in it to select the function version to use after \token{range(} in some context, using the English words \token{start} and \token{end}. 
Specifically, we take all single-argument \token{range(end)} calls in the MBPP dataset ($N=159$) and attempt to expand it into \token{range(0, end)}.
We use a similar intervention method, except simply using the unembeddings of trigger tokens instead of intermediate LM hidden states. We use as trigger tokens (\token{start} -- \token{end}) and add the intervention vector to the hidden states corresponding to the open parenthesis \token{(} at an intermediate layer (layer 17).  For all these \token{range} calls in the dataset, we let Llama-2\footnote{We do not consider Llama-3 in this case because its default behavior usually generates \token{range(0, end)} in the first place, and it is unclear how to steer from \token{range(0, end)} to \token{range(end)}.} generate without and with intervention.
Figure~\ref{fig:code-intervention} shows that, with increasing intervention strength, more instances are successfully steered to \token{range(0, end)}, up to 67\%.

\paragraph{Formal semantics.}
Unlike in the cases above where we modify the surface continuation by manipulating some underlying concept structure, a natural language prefix usually licenses one single possible thematic role to follow. So we do not perform an intervention experiment here.

\label{sec:intervention-visionlanguage}
\paragraph{Visual input.} We show that we can steer the output of Chameleon by intervening on the image patches using language tokens and analyze how this affects the textual output.
Focusing on the color setup, if the representation for a color is similar between visual and language inputs, we hypothesize that we can \emph{replace} the image hidden states corresponding to one pure color image patch with the unembedding of the language token for another color, and mislead the model to ``perceive'' the new color when asked about the image color. Note that replacing the hidden state is a more invasive intervention than addition.
However, there is a confounder: the intervened word may lexically bias the model to generate the same word, without performing reasoning that incorporates the new color, because the desired answer after intervening is equivalent to the intervention word itself (unlike in the previous cases). To control for this, we show two colors in one image and only intervene at the positions corresponding to one color: if the intervention unconditionally and lexically biases the generation to the new color, this effect would (incorrectly) affect both colors.\footnote{We tested settings that require more sophisticated reasoning such as asking for a country flag with the two colors, or asking about spatial relationships of the colors. They seem to be beyond the capability of Chameleon-7B---even without interventions, the model cannot answer the questions correctly.}

We consider all color pairs using the same colors as in \S\ref{sec:identify-visionlanguage}: red, green, blue, and black,\footnote{We tried other colors but Chameleon-7B, without interventions, cannot recognize those colors reliably} and picking one color in the pair and intervene to a new third color ($N=48$). As the intervention, we start from a layer $\ell$ and replace all hidden states at and after $\ell$ to be the unembedding of the new color minus the old color. We ask the model what the two colors in the image are, and only consider the intervention successful if the model answers both the new color and the other unintervened color correctly. Figure~\ref{fig:color-intervention} shows the success rate across all $\ell$: it gets as high as above 80\%.\footnote{One may argue this is conceptually similar to a half-language half-image input. There are many distinctions: most importantly, a half-image is not processable by Chameleon and severely goes out of its training distribution, since it only ever processes images of size exactly $512 \times 512$. Other distinctions include: the presence of a special token marking the beginning of the image; our intervention repeats the new color token, once for each patch, rather than just one; and the token representation is held constant across layers rather than evolving; etc.} We highlight that, for both this experiment and the earlier ones in this section, the interventions are not even necessarily guaranteed to lead sensible outputs, let alone correct ones.

\paragraph{Audio.} We perform a similar intervention with SALMONN, with the same desideratum that the question-answering process should require some reasoning rather than outputting the intervened token as-is. We consider 1000 animal sounds in the VGGSound dataset, specifically only single-word animals, and ask \token{Is this animal a mammal?} We intervene both on mammal sounds with a random non-mammal word (and expecting the model to be more likely to reply with \token{No}) and vice versa.
We perform the invention similarly to the code case, adding the unembedding difference between the new trigger word and the original animal name, scaled by a constant, at layer 13.
We measure the probability of the \token{Yes} token and the \token{No} token and compute the normalized \token{Yes} probability.
Figure~\ref{fig:audio-intervention} visualizes the two cases across multiple intervention strengths.
As the strength increases, the model is more likely to predict in the steered direction, again demonstrating cross-data-type intervention effectiveness.

\paragraph{Note on the intervention methods used.}
In the majority of our experiments, we add an intervention vector to a particular layer at a particular input position, and the intervention vector is computed using the difference between the unembeddings of two trigger tokens in another data type, scaled by a constant. This is simple and naturally follows the results of our logit lens experiments. However, this may not be convincing for data types highly parallel to English. For example, for the multilingual intervention where we use the difference between the unembeddings of, e.g., \token{Good} and \token{Bad}, the intervention may succeed only because this difference is similar to the unembedding difference between \token{Bueno} (trans. ``Good'') and \token{Malo} (trans. ``Bad'') due to cross-lingual embedding alignment~\citepia{mikolov2013exploiting}.\footnote{Note that this is less problematic for other data types such as code, where it is unlikely that the unembedding difference between \token{end} and \token{start} is similar to that between the actual surface form end argument and \token{0}.} So instead, as the intervention vector, we use not the unembedding difference but the difference between intermediate model hidden states, and they \emph{are} expected to be the same across modalities as our hypothesis predicts. We do this for multilingual intervention and arithmetic intervention. The former, using ActAdd, further differs by adding the vector not at a targeted position but in the beginning of the sentence. In addition to it being an established intervention method, we used it because in sentiment steering, there is no targeted position of interest (unlike for arithmetic), but we want to steer the sentence representation globally. Finally, we use a more invasive intervention for visual inputs where we do not just add, but replace, hidden representations because we found it to work well, providing a stronger result.

\section{Discussion}

Our experiments across diverse languages and modalities give empirical support to the semantic hub hypothesis; language models seem to make efficient use of model capacity and learn to represent semantically similar inputs from different modalities near one another. Such a semantic hub has been hypothesized to exist in the human brain~\citep{patterson2007you,patterson2016hub,ralph2017neural}. Inasmuch as both the human brain and language models are resource-constrained information processing systems, it is perhaps unsurprising that both systems make use of a semantic hub. This also supports prior findings that multilingual models usually reason best in English~\citep{shi2023language}.

We note however that this dominant-language-reliance may not always be a desirable property of language models. \citet{wendler-etal-2024-llamas} conjectured that language models might inherit biases present in the training data of the dominant language; if true, an extreme implementation of this strategy could force an alignment of ideology across languages and potentially harm inclusivity.
Similarly, internalizing arithmetic expressions in natural language may not be ideal as a part of the ``algorithm'' with which a language model implements arithmetic, considering their structural differences and the vast linguistic variation in number expressions in their composition (c.f. Danish ``\emph{tooghalvfems}''; trans. ninety-two, but compositionally representing \texttt{2+(5-0.5)$\times$20}) or even bases (c.f. the base-23 numeral system in the Kalam language~\citep{laycock}).
In general, an over-reliance on  the dominant data type could be detrimental.

\section{Related Work}

\paragraph{Representation alignment between separately trained models.}
A long line of work has investigated the representations of separately trained mono-data-type models, and showed that they can be aligned through a transformation.
In the multilingual case, it has been found that separately trained word embeddings for different languages can be aligned~\citepia{mikolov2013exploiting,smith2017offline,Cao2020Multilingual}.
Similarly, prior work has shown that visual representations and text representations from different models can be mapped together~\citepia{merullo2022linearly,10.5555/3618408.3619119,Maniparambil_2024_CVPR}.
\citet{huh2024platonic} argued that these are possible because the different data modalities are projections of the same underlying reality.
\citet{lu2021fpt} showed that language-only models can be minimally finetuned to achieve high performance in other modalities.
Our work, in contrast, looks at a \emph{single} static model that processes multiple input data types, and finds that the resulting representations align, without needing a transformation.
Concurrent work~\citep{luo2024taskvectorscrossmodal} found similar alignments of ``task vectors''~\citep{ilharco2022editing,hendel-etal-2023-context} in vision-language models.

\paragraph{Representation evolution throughout layers.}
Much past work has analyzed the representation evolution throughout transformer layers, inspecting how it affects reasoning~\citep{yang-etal-2024-large-language-models}, factuality~\citep{chuang2024dola}, knowledge~\citep{jin2024exploring}, etc. From another angle, work on layer pruning and early exiting also speaks to the representation dynamics across layers~\citepia{gromov2024unreasonableineffectivenessdeeperlayers,sanyal2024pretrainingsmallbaselms}.
More mechanistically, \citet{merullo-etal-2024-language}, \citet{todd2024function}, \citet{hendel-etal-2023-context}, \emph{i.a.}, more precisely characterized the representation changes algorithmically.

\paragraph{Inspecting model hidden states.}
We adopted the logit lens for its simplicity which brings few confounders. However, alternatives exist, usually requiring some training~\citepia{belrose2023elicitinglatentpredictionstransformers,ghandeharioun2024patchscopesunifyingframeworkinspecting,templeton2024scaling}. They allow for more expressive explanations, though at the risk of overfitting. Similar methods have been developed for other modalities, such as \citet{toker-etal-2024-diffusion}. Testing our hypothesis using these methods would be valuable future work.
\section{Conclusion}

This work proposed and investigated the semantic hub hypothesis, which posits that language models represent semantically similar inputs from distinct modalities near one another in their intermediate layers. We find evidence of this phenomenon across multiple language models and data types, and further observe that intervening in this space through the model's dominant language (usually English) leads to predictable  model behavior changes.

\ificlrfinal
\subsubsection*{Acknowledgments}
We thank Alex Gu, Alexis Ross, Alisa Liu, Aryaman Arora, Asma Ghandeharioun, Cedegao E. Zhang, Freda Shi, Han Guo, Jack Merullo, Jonathan May, Linlu Qiu, Mor Geva, Naman Jain, Ruochen Zhang, Sarah Wiegreffe, Shushan Arakelyan, and Yung-Sung Chuang for discussions and help at various stages of this project.
Figure~\ref{fig:overview} uses icons from \url{flaticon.com}. This study was supported by funds from MIT-IBM Watson AI Lab.
\fi

\bibliography{iclr2024_conference}

\begin{thebibliography}{85}
\providecommand{\natexlab}[1]{#1}
\providecommand{\url}[1]{\texttt{#1}}
\expandafter\ifx\csname urlstyle\endcsname\relax
  \providecommand{\doi}[1]{doi: #1}\else
  \providecommand{\doi}{doi: \begingroup \urlstyle{rm}\Url}\fi

\bibitem[Alabi et~al.(2024)Alabi, Mosbach, Eyal, Klakow, and Geva]{alabi-etal-2024-hidden}
Jesujoba Alabi, Marius Mosbach, Matan Eyal, Dietrich Klakow, and Mor Geva.
\newblock The hidden space of transformer language adapters.
\newblock In \emph{Proceedings of the Annual Meeting of the Association for Computational Linguistics (ACL)}, 2024.
\newblock URL \url{https://aclanthology.org/2024.acl-long.356}.

\bibitem[Artetxe et~al.(2017)Artetxe, Labaka, and Agirre]{artetxe-etal-2017-learning}
Mikel Artetxe, Gorka Labaka, and Eneko Agirre.
\newblock Learning bilingual word embeddings with (almost) no bilingual data.
\newblock In \emph{Proceedings of Annual Meeting of the Association for Computational Linguistics (ACL)}, 2017.
\newblock URL \url{https://aclanthology.org/P17-1042}.

\bibitem[Austin et~al.(2021)Austin, Odena, Nye, Bosma, Michalewski, Dohan, Jiang, Cai, Terry, Le, et~al.]{austin2021program}
Jacob Austin, Augustus Odena, Maxwell Nye, Maarten Bosma, Henryk Michalewski, David Dohan, Ellen Jiang, Carrie Cai, Michael Terry, Quoc Le, et~al.
\newblock Program synthesis with large language models.
\newblock \emph{arXiv preprint}, 2021.
\newblock URL \url{https://arxiv.org/abs/2108.07732}.

\bibitem[Belrose et~al.(2023)Belrose, Furman, Smith, Halawi, Ostrovsky, McKinney, Biderman, and Steinhardt]{belrose2023elicitinglatentpredictionstransformers}
Nora Belrose, Zach Furman, Logan Smith, Danny Halawi, Igor Ostrovsky, Lev McKinney, Stella Biderman, and Jacob Steinhardt.
\newblock Eliciting latent predictions from transformers with the tuned lens.
\newblock \emph{arXiv preprint}, 2023.
\newblock URL \url{https://arxiv.org/abs/2303.08112}.

\bibitem[Beyer et~al.(1999)Beyer, Goldstein, Ramakrishnan, and Shaft]{beyer1999}
Kevin~S. Beyer, Jonathan Goldstein, Raghu Ramakrishnan, and Uri Shaft.
\newblock When is ''nearest neighbor'' meaningful?
\newblock In \emph{Proceedings of the 7th International Conference on Database Theory}, ICDT '99, pp.\  217–235, Berlin, Heidelberg, 1999. Springer-Verlag.
\newblock ISBN 3540654526.

\bibitem[BigScience(2023)]{workshop2023bloom176bparameteropenaccessmultilingual}
BigScience.
\newblock Bloom: A 176b-parameter open-access multilingual language model.
\newblock \emph{arXiv Preprint}, 2023.
\newblock URL \url{https://arxiv.org/abs/2211.05100}.

\bibitem[Cao et~al.(2020)Cao, Kitaev, and Klein]{Cao2020Multilingual}
Steven Cao, Nikita Kitaev, and Dan Klein.
\newblock Multilingual alignment of contextual word representations.
\newblock In \emph{Proceedings of the International Conference on Learning Representations (ICLR)}, 2020.
\newblock URL \url{https://openreview.net/forum?id=r1xCMyBtPS}.

\bibitem[Chameleon-Team(2024)]{Chameleon_Team_Chameleon_Mixed-Modal_Early-Fusion_2024}
Chameleon-Team.
\newblock Chameleon: Mixed-modal early-fusion foundation models.
\newblock \emph{arXiv preprint}, 2024.
\newblock URL \url{https://arXiv.org/abs/2405.09818}.

\bibitem[Chan et~al.(2022)Chan, Garriga-Alonso, Goldwosky-Dill, Greenblatt, Nitishinskaya, Radhakrishnan, Shlegeris, and Thomas]{chan2022causal}
Lawrence Chan, Adrià Garriga-Alonso, Nicholas Goldwosky-Dill, Ryan Greenblatt, Jenny Nitishinskaya, Ansh Radhakrishnan, Buck Shlegeris, and Nate Thomas.
\newblock Causal scrubbing, a method for rigorously testing interpretability hypotheses.
\newblock \emph{AI Alignment Forum}, 2022.
\newblock URL \url{https://www.alignmentforum.org/posts/JvZhhzycHu2Yd57RN/causal-scrubbing-a-method-for-rigorously-testing}.

\bibitem[Chen et~al.(2020)Chen, Xie, Vedaldi, and Zisserman]{Chen20VGGS}
Honglie Chen, Weidi Xie, Andrea Vedaldi, and Andrew Zisserman.
\newblock Vggsound: A large-scale audio-visual dataset.
\newblock In \emph{Proceedings of the International Conference on Acoustics, Speech, and Signal Processing (ICASSP)}, 2020.
\newblock URL \url{https://ieeexplore.ieee.org/document/9053174}.

\bibitem[Chen et~al.(2016)Chen, Krug, and Strassel]{gale2016}
Song Chen, Gary Krug, and Stephanie Strassel.
\newblock Gale phase 3 and 4 chinese newswire parallel text.
\newblock \emph{Linguistic Data Consortium}, 2016.
\newblock URL \url{https://catalog.ldc.upenn.edu/LDC2016T25}.

\bibitem[Chuang et~al.(2024)Chuang, Xie, Luo, Kim, Glass, and He]{chuang2024dola}
Yung-Sung Chuang, Yujia Xie, Hongyin Luo, Yoon Kim, James~R. Glass, and Pengcheng He.
\newblock Dola: Decoding by contrasting layers improves factuality in large language models.
\newblock In \emph{Proceedings of the International Conference on Learning Representations (ICLR)}, 2024.
\newblock URL \url{https://openreview.net/forum?id=Th6NyL07na}.

\bibitem[Conneau et~al.(2018)Conneau, Lample, Ranzato, Denoyer, and J{\'e}gou]{conneau2017word}
Alexis Conneau, Guillaume Lample, Marc'Aurelio Ranzato, Ludovic Denoyer, and Herv{\'e} J{\'e}gou.
\newblock Word translation without parallel data.
\newblock In \emph{Proceedings of International Conference on Learning Representations (ICLR)}, 2018.
\newblock URL \url{https://arxiv.org/abs/1710.04087}.

\bibitem[Conneau et~al.(2020)Conneau, Khandelwal, Goyal, Chaudhary, Wenzek, Guzm{\'a}n, Grave, Ott, Zettlemoyer, and Stoyanov]{conneau-etal-2020-unsupervised}
Alexis Conneau, Kartikay Khandelwal, Naman Goyal, Vishrav Chaudhary, Guillaume Wenzek, Francisco Guzm{\'a}n, Edouard Grave, Myle Ott, Luke Zettlemoyer, and Veselin Stoyanov.
\newblock Unsupervised cross-lingual representation learning at scale.
\newblock In \emph{Proceedings of the Annual Meeting of the Association for Computational Linguistics (ACL)}, 2020.
\newblock URL \url{https://aclanthology.org/2020.acl-main.747}.

\bibitem[D{\'i}az-Galiano et~al.(2018)D{\'i}az-Galiano, Mart{\'i}nez-C{\'a}mara, Cumbreras, Vega, and Villena-Rom{\'a}n]{DazGaliano2018TheDO}
Manuel~Carlos D{\'i}az-Galiano, Eugenio Mart{\'i}nez-C{\'a}mara, Miguel {\'A}ngel~Garc{\'i}a Cumbreras, Manuel~Garc{\'i}a Vega, and Julio Villena-Rom{\'a}n.
\newblock The democratization of deep learning in tass 2017.
\newblock \emph{Proces. del Leng. Natural}, 2018.
\newblock URL \url{https://api.semanticscholar.org/CorpusID:13667878}.

\bibitem[Dumas et~al.(2024)Dumas, Veselovsky, Monea, West, and Wendler]{dumas2024how}
Cl{\'e}ment Dumas, Veniamin Veselovsky, Giovanni Monea, Robert West, and Chris Wendler.
\newblock How do llamas process multilingual text? a latent exploration through activation patching.
\newblock In \emph{Proceedings of the ICML Workshop on Mechanistic Interpretability}, 2024.
\newblock URL \url{https://openreview.net/forum?id=0ku2hIm4BS}.

\bibitem[Elazar et~al.(2021)Elazar, Ravfogel, Jacovi, and Goldberg]{elazar2021amnesic}
Yanai Elazar, Shauli Ravfogel, Alon Jacovi, and Yoav Goldberg.
\newblock {Amnesic Probing: Behavioral Explanation with Amnesic Counterfactuals}.
\newblock \emph{Transactions of the Association for Computational Linguistics (TACL)}, 2021.
\newblock URL \url{https://doi.org/10.1162/tacl\_a\_00359}.

\bibitem[Fillmore(1968)]{fillmore}
Charles~J. Fillmore.
\newblock The case for case.
\newblock In Emmon Bach and Robert~T. Harms (eds.), \emph{Universals in Linguistic Theory}, pp.\  0--88. Holt, Rinehart and Winston, New York, 1968.

\bibitem[Gemini-Team(2024)]{geminiteam2024geminifamilyhighlycapable}
Gemini-Team.
\newblock Gemini: A family of highly capable multimodal models.
\newblock \emph{arXiv preprint}, 2024.
\newblock URL \url{https://arxiv.org/abs/2312.11805}.

\bibitem[Ghandeharioun et~al.(2024)Ghandeharioun, Caciularu, Pearce, Dixon, and Geva]{ghandeharioun2024patchscopesunifyingframeworkinspecting}
Asma Ghandeharioun, Avi Caciularu, Adam Pearce, Lucas Dixon, and Mor Geva.
\newblock Patchscopes: A unifying framework for inspecting hidden representations of language models.
\newblock \emph{arXiv preprint}, 2024.
\newblock URL \url{https://arxiv.org/abs/2401.06102}.

\bibitem[Gong et~al.(2023)Gong, Liu, Luo, Karlinsky, and Glass]{jasu}
Yuan Gong, Alexander~H. Liu, Hongyin Luo, Leonid Karlinsky, and James Glass.
\newblock Joint audio and speech understanding.
\newblock In \emph{Proceedings of the IEEE Automatic Speech Recognition and Understanding Workshop (ASRU)}, 2023.
\newblock URL \url{https://arxiv.org/abs/2309.14405}.

\bibitem[Gong et~al.(2024)Gong, Luo, Liu, Karlinsky, and Glass]{gong2024listen}
Yuan Gong, Hongyin Luo, Alexander~H. Liu, Leonid Karlinsky, and James~R. Glass.
\newblock Listen, think, and understand.
\newblock In \emph{Proceedings of the International Conference on Learning Representations (ICLR)}, 2024.
\newblock URL \url{https://openreview.net/forum?id=nBZBPXdJlC}.

\bibitem[Gromov et~al.(2024)Gromov, Tirumala, Shapourian, Glorioso, and Roberts]{gromov2024unreasonableineffectivenessdeeperlayers}
Andrey Gromov, Kushal Tirumala, Hassan Shapourian, Paolo Glorioso, and Daniel~A. Roberts.
\newblock The unreasonable ineffectiveness of the deeper layers.
\newblock \emph{arXiv preprint}, 2024.
\newblock URL \url{https://arxiv.org/abs/2403.17887}.

\bibitem[Hendel et~al.(2023)Hendel, Geva, and Globerson]{hendel-etal-2023-context}
Roee Hendel, Mor Geva, and Amir Globerson.
\newblock In-context learning creates task vectors.
\newblock In \emph{Findings of the Association for Computational Linguistics: EMNLP}, December 2023.
\newblock URL \url{https://aclanthology.org/2023.findings-emnlp.624}.

\bibitem[Honnibal \& Montani(2017)Honnibal and Montani]{spacy2}
Matthew Honnibal and Ines Montani.
\newblock {spaCy 2}: Natural language understanding with {B}loom embeddings, convolutional neural networks and incremental parsing.
\newblock To appear, 2017.
\newblock URL \url{https://spacy.io}.

\bibitem[Hua et~al.(2024)Hua, Yun, and Pavlick]{hua-etal-2024-mothello}
Tianze Hua, Tian Yun, and Ellie Pavlick.
\newblock m{O}thello: When do cross-lingual representation alignment and cross-lingual transfer emerge in multilingual models?
\newblock In \emph{Findings of the Association for Computational Linguistics: NAACL 2024}, 2024.
\newblock URL \url{https://aclanthology.org/2024.findings-naacl.103}.

\bibitem[Huh et~al.(2024)Huh, Cheung, Wang, and Isola]{huh2024platonic}
Minyoung Huh, Brian Cheung, Tongzhou Wang, and Phillip Isola.
\newblock The platonic representation hypothesis.
\newblock In \emph{Proceedings of Machine Learning Research (ICML)}, 2024.
\newblock URL \url{https://proceedings.mlr.press/v235/huh24a.html}.

\bibitem[Ilharco et~al.(2021)Ilharco, Zellers, Farhadi, and Hajishirzi]{ilharco2020probing}
Gabriel Ilharco, Rowan Zellers, Ali Farhadi, and Hannaneh Hajishirzi.
\newblock Probing contextual language models for common ground with visual representations.
\newblock In \emph{Proceedings of the Conference of the North {A}merican Chapter of the Association for Computational Linguistics: Human Language Technologies (NAACL-HLT)}, 2021.
\newblock URL \url{https://aclanthology.org/2021.naacl-main.422}.

\bibitem[Ilharco et~al.(2022)Ilharco, Ribeiro, Wortsman, Gururangan, Schmidt, Hajishirzi, and Farhadi]{ilharco2022editing}
Gabriel Ilharco, Marco~Tulio Ribeiro, Mitchell Wortsman, Suchin Gururangan, Ludwig Schmidt, Hannaneh Hajishirzi, and Ali Farhadi.
\newblock Editing models with task arithmetic.
\newblock \emph{arXiv preprint arXiv:2212.04089}, 2022.

\bibitem[Jackendoff(1974)]{jackendoff1974semantic}
R.S. Jackendoff.
\newblock \emph{Semantic Interpretation in Generative Grammar}.
\newblock Studies in linguistics series. MIT Press, 1974.
\newblock ISBN 9780262600071.

\bibitem[Jin et~al.(2024)Jin, Yu, Huang, Zeng, Wang, Hua, Zhao, Mei, Meng, Ding, et~al.]{jin2024exploring}
Mingyu Jin, Qinkai Yu, Jingyuan Huang, Qingcheng Zeng, Zhenting Wang, Wenyue Hua, Haiyan Zhao, Kai Mei, Yanda Meng, Kaize Ding, et~al.
\newblock Exploring concept depth: How large language models acquire knowledge at different layers?
\newblock \emph{arXiv preprint}, 2024.
\newblock URL \url{https://arxiv.org/abs/2404.07066}.

\bibitem[Keung et~al.(2020)Keung, Lu, Szarvas, and Smith]{keung-etal-2020-multilingual}
Phillip Keung, Yichao Lu, Gy{\"o}rgy Szarvas, and Noah~A. Smith.
\newblock The multilingual {A}mazon reviews corpus.
\newblock In \emph{Proceedings of the Conference on Empirical Methods in Natural Language Processing (EMNLP)}, 2020.
\newblock URL \url{https://aclanthology.org/2020.emnlp-main.369}.

\bibitem[Kim \& Linzen(2020)Kim and Linzen]{kim-linzen-2020-cogs}
Najoung Kim and Tal Linzen.
\newblock {COGS}: A compositional generalization challenge based on semantic interpretation.
\newblock In \emph{Proceedings of the Conference on Empirical Methods in Natural Language Processing (EMNLP)}, 2020.
\newblock URL \url{https://aclanthology.org/2020.emnlp-main.731}.

\bibitem[Koh et~al.(2023)Koh, Salakhutdinov, and Fried]{10.5555/3618408.3619119}
Jing~Yu Koh, Ruslan Salakhutdinov, and Daniel Fried.
\newblock Grounding language models to images for multimodal inputs and outputs.
\newblock In \emph{Proceedings of the International Conference on Machine Learning (ICLR)}, 2023.
\newblock URL \url{https://arxiv.org/abs/2301.13823}.

\bibitem[Laycock(1975)]{laycock}
Donald~C Laycock.
\newblock \emph{Observations on Number Systems and Semantics}.
\newblock Pacific Linguistics, 1975.

\bibitem[Li et~al.(2021)Li, Nye, and Andreas]{li-etal-2021-implicit}
Belinda~Z. Li, Maxwell Nye, and Jacob Andreas.
\newblock Implicit representations of meaning in neural language models.
\newblock In \emph{Proceedings of the Annual Meeting of the Association for Computational Linguistics and the International Joint Conference on Natural Language Processing (ACL-IJCNLP)}, 2021.
\newblock URL \url{https://aclanthology.org/2021.acl-long.143}.

\bibitem[Li et~al.(2023)Li, Kementchedjhieva, and S{\o}gaard]{li2023implications}
Jiaang Li, Yova Kementchedjhieva, and Anders S{\o}gaard.
\newblock Implications of the convergence of language and vision model geometries.
\newblock \emph{arXiv preprint}, 2023.
\newblock URL \url{https://arxiv.org/abs/2302.06555}.

\bibitem[Lin et~al.(2014)Lin, Maire, Belongie, Hays, Perona, Ramanan, Dollar, and Zitnick]{coco}
Tsung-Yi Lin, Michael Maire, Serge Belongie, James Hays, Pietro Perona, Deva Ramanan, Piotr Dollar, and Larry Zitnick.
\newblock Microsoft coco: Common objects in context.
\newblock In \emph{Proceedings of the European Conference on Computer Vision (ECCV)}, 2014.
\newblock URL \url{https://www.microsoft.com/en-us/research/publication/microsoft-coco-common-objects-in-context/}.

\bibitem[Liu et~al.(2023)Liu, Li, Wu, and Lee]{liu2023llava}
Haotian Liu, Chunyuan Li, Qingyang Wu, and Yong~Jae Lee.
\newblock Visual instruction tuning.
\newblock In \emph{Advances in Neural Information Processing Systems (NeurIPS)}, 2023.
\newblock URL \url{https://arxiv.org/abs/2304.08485}.

\bibitem[Llama-3-Team(2024)]{dubey2024llama3herdmodels}
The Llama-3-Team.
\newblock The llama 3 herd of models.
\newblock \emph{arXiv Preprint}, 2024.
\newblock URL \url{https://arxiv.org/abs/2407.21783}.

\bibitem[Lu et~al.(2023)Lu, Clark, Zellers, Mottaghi, and Kembhavi]{uio}
Jiasen Lu, Christopher Clark, Rowan Zellers, Roozbeh Mottaghi, and Aniruddha Kembhavi.
\newblock {UNIFIED}-{IO}: A unified model for vision, language, and multi-modal tasks.
\newblock In \emph{Proceedings of the International Conference on Learning Representations (ICLR)}, 2023.
\newblock URL \url{https://openreview.net/forum?id=E01k9048soZ}.

\bibitem[Lu et~al.(2024)Lu, Clark, Lee, Zhang, Khosla, Marten, Hoiem, and Kembhavi]{uio2}
Jiasen Lu, Christopher Clark, Sangho Lee, Zichen Zhang, Savya Khosla, Ryan Marten, Derek Hoiem, and Aniruddha Kembhavi.
\newblock {Unified-IO} 2: Scaling autoregressive multimodal models with vision language audio and action.
\newblock In \emph{Proceedings of the IEEE/CVF Conference on Computer Vision and Pattern Recognition (CVPR)}, 2024.
\newblock URL \url{https://arxiv.org/abs/2312.17172}.

\bibitem[Lu et~al.(2021)Lu, Grover, Abbeel, and Mordatch]{lu2021fpt}
Kevin Lu, Aditya Grover, Pieter Abbeel, and Igor Mordatch.
\newblock Pretrained transformers as universal computation engines.
\newblock \emph{arXiv preprint arXiv:2103.05247}, 2021.

\bibitem[Luo et~al.(2024)Luo, Darrell, and Bar]{luo2024taskvectorscrossmodal}
Grace Luo, Trevor Darrell, and Amir Bar.
\newblock Task vectors are cross-modal, 2024.
\newblock URL \url{https://arxiv.org/abs/2410.22330}.

\bibitem[Maniparambil et~al.(2024)Maniparambil, Akshulakov, Djilali, El~Amine~Seddik, Narayan, Mangalam, and O'Connor]{Maniparambil_2024_CVPR}
Mayug Maniparambil, Raiymbek Akshulakov, Yasser Abdelaziz~Dahou Djilali, Mohamed El~Amine~Seddik, Sanath Narayan, Karttikeya Mangalam, and Noel~E. O'Connor.
\newblock Do vision and language encoders represent the world similarly?
\newblock In \emph{Proceedings of the IEEE/CVF Conference on Computer Vision and Pattern Recognition (CVPR)}, 2024.
\newblock URL \url{https://arxiv.org/abs/2401.05224}.

\bibitem[{MDBG}(2024)]{mdbg_cc_cedict}
{MDBG}.
\newblock {MDBG} {Chinese}-{English} dictionary ({CC-CEDICT}).
\newblock MBDG, 2024.
\newblock URL \url{https://www.mdbg.net/chinese/dictionary?page=cc-cedict}.
\newblock Downloaded: 2024-09-25.

\bibitem[Merullo et~al.(2022)Merullo, Castricato, Eickhoff, and Pavlick]{merullo2022linearly}
Jack Merullo, Louis Castricato, Carsten Eickhoff, and Ellie Pavlick.
\newblock Linearly mapping from image to text space.
\newblock \emph{arXiv preprint}, 2022.
\newblock URL \url{https://arxiv.org/abs/2209.15162}.

\bibitem[Merullo et~al.(2024)Merullo, Eickhoff, and Pavlick]{merullo-etal-2024-language}
Jack Merullo, Carsten Eickhoff, and Ellie Pavlick.
\newblock Language models implement simple {W}ord2{V}ec-style vector arithmetic.
\newblock In \emph{Proceedings of the Conference of the North American Chapter of the Association for Computational Linguistics: Human Language Technologies (NAACL-HLT)}, 2024.
\newblock URL \url{https://aclanthology.org/2024.naacl-long.281}.

\bibitem[Mikolov et~al.(2013)Mikolov, Le, and Sutskever]{mikolov2013exploiting}
Tomas Mikolov, Quoc~V Le, and Ilya Sutskever.
\newblock Exploiting similarities among languages for machine translation.
\newblock \emph{arXiv preprint}, 2013.
\newblock URL \url{https://arxiv.org/abs/1309.4168}.

\bibitem[Morris et~al.(2023)Morris, Kuleshov, Shmatikov, and Rush]{morris-etal-2023-text}
John Morris, Volodymyr Kuleshov, Vitaly Shmatikov, and Alexander Rush.
\newblock Text embeddings reveal (almost) as much as text.
\newblock In \emph{Proceedings of the Conference on Empirical Methods in Natural Language Processing (EMNLP)}, 2023.
\newblock URL \url{https://aclanthology.org/2023.emnlp-main.765}.

\bibitem[Ngo \& Kim(2024)Ngo and Kim]{ngo2024language}
Jerry Ngo and Yoon Kim.
\newblock What do language models hear? probing for auditory representations in language models.
\newblock In \emph{Proceedings of Annual Meeting of the Association for Computational Linguistics (ACL)}, 2024.
\newblock URL \url{https://arxiv.org/abs/2402.16998}.

\bibitem[nostalgebraist(2020)]{logit-lens}
nostalgebraist.
\newblock Interpreting {GPT}: the logit lens.
\newblock LessWrong, 2020.
\newblock URL \url{https://www.lesswrong.com/posts/AcKRB8wDpdaN6v6ru/interpreting-gpt-the-logit-lens}.

\bibitem[Parsons(1990)]{Parsons1990-PAREIT}
Terence Parsons.
\newblock \emph{Events in the Semantics of English: A Study in Subatomic Semantics}.
\newblock MIT Press, 1990.

\bibitem[Patterson \& Ralph(2016)Patterson and Ralph]{patterson2016hub}
Karalyn Patterson and Matthew A~Lambon Ralph.
\newblock The hub-and-spoke hypothesis of semantic memory.
\newblock In \emph{Neurobiology of language}, pp.\  765--775. Elsevier, 2016.

\bibitem[Patterson et~al.(2007)Patterson, Nestor, and Rogers]{patterson2007you}
Karalyn Patterson, Peter~J Nestor, and Timothy~T Rogers.
\newblock Where do you know what you know? the representation of semantic knowledge in the human brain.
\newblock \emph{Nature reviews neuroscience}, 8\penalty0 (12):\penalty0 976--987, 2007.

\bibitem[Radford et~al.(2021)Radford, Kim, Hallacy, Ramesh, Goh, Agarwal, Sastry, Askell, Mishkin, Clark, Krueger, and Sutskever]{clip}
Alec Radford, Jong~Wook Kim, Chris Hallacy, Aditya Ramesh, Gabriel Goh, Sandhini Agarwal, Girish Sastry, Amanda Askell, Pamela Mishkin, Jack Clark, Gretchen Krueger, and Ilya Sutskever.
\newblock Learning transferable visual models from natural language supervision.
\newblock In Marina Meila and Tong Zhang (eds.), \emph{Proceedings of the 38th International Conference on Machine Learning}, volume 139 of \emph{Proceedings of Machine Learning Research}, pp.\  8748--8763. PMLR, 18--24 Jul 2021.
\newblock URL \url{https://proceedings.mlr.press/v139/radford21a.html}.

\bibitem[Ralph et~al.(2017)Ralph, Jefferies, Patterson, and Rogers]{ralph2017neural}
Matthew A~Lambon Ralph, Elizabeth Jefferies, Karalyn Patterson, and Timothy~T Rogers.
\newblock The neural and computational bases of semantic cognition.
\newblock \emph{Nature reviews neuroscience}, 18\penalty0 (1):\penalty0 42--55, 2017.

\bibitem[Ravichander et~al.(2021)Ravichander, Belinkov, and Hovy]{ravichander-etal-2021-probing}
Abhilasha Ravichander, Yonatan Belinkov, and Eduard Hovy.
\newblock Probing the probing paradigm: Does probing accuracy entail task relevance?
\newblock In \emph{Proceedings of the Conference of the European Chapter of the Association for Computational Linguistics (EACL)}, 2021.
\newblock URL \url{https://aclanthology.org/2021.eacl-main.295}.

\bibitem[Rimsky et~al.(2024)Rimsky, Gabrieli, Schulz, Tong, Hubinger, and Turner]{rimsky-etal-2024-steering}
Nina Rimsky, Nick Gabrieli, Julian Schulz, Meg Tong, Evan Hubinger, and Alexander Turner.
\newblock Steering llama 2 via contrastive activation addition.
\newblock In \emph{Proceedings of the 62nd Annual Meeting of the Association for Computational Linguistics (ACL)}, 2024.
\newblock URL \url{https://aclanthology.org/2024.acl-long.828}.

\bibitem[Sanh et~al.(2020)Sanh, Debut, Chaumond, and Wolf]{sanh2020distilbertdistilledversionbert}
Victor Sanh, Lysandre Debut, Julien Chaumond, and Thomas Wolf.
\newblock Distilbert, a distilled version of bert: smaller, faster, cheaper and lighter.
\newblock \emph{arXiv preprint}, 2020.
\newblock URL \url{https://arxiv.org/abs/1910.01108}.

\bibitem[Sanyal et~al.(2024)Sanyal, Sanghavi, and Dimakis]{sanyal2024pretrainingsmallbaselms}
Sunny Sanyal, Sujay Sanghavi, and Alexandros~G. Dimakis.
\newblock Pre-training small base lms with fewer tokens.
\newblock \emph{arXiv preprint}, 2024.
\newblock URL \url{https://arxiv.org/abs/2404.08634}.

\bibitem[Schuster et~al.(2019)Schuster, Ram, Barzilay, and Globerson]{schuster-etal-2019-cross}
Tal Schuster, Ori Ram, Regina Barzilay, and Amir Globerson.
\newblock Cross-lingual alignment of contextual word embeddings, with applications to zero-shot dependency parsing.
\newblock In \emph{Proceedings of the Conference of the North {A}merican Chapter of the Association for Computational Linguistics: Human Language Technologies (NAACL-HLT)}, 2019.
\newblock URL \url{https://aclanthology.org/N19-1162}.

\bibitem[Shi et~al.(2023)Shi, Suzgun, Freitag, Wang, Srivats, Vosoughi, Chung, Tay, Ruder, Zhou, Das, and Wei]{shi2023language}
Freda Shi, Mirac Suzgun, Markus Freitag, Xuezhi Wang, Suraj Srivats, Soroush Vosoughi, Hyung~Won Chung, Yi~Tay, Sebastian Ruder, Denny Zhou, Dipanjan Das, and Jason Wei.
\newblock Language models are multilingual chain-of-thought reasoners.
\newblock In \emph{The Eleventh International Conference on Learning Representations}, 2023.
\newblock URL \url{https://openreview.net/forum?id=fR3wGCk-IXp}.

\bibitem[Smith et~al.(2017)Smith, Turban, Hamblin, and Hammerla]{smith2017offline}
Samuel~L. Smith, David H.~P. Turban, Steven Hamblin, and Nils~Y. Hammerla.
\newblock Offline bilingual word vectors, orthogonal transformations and the inverted softmax.
\newblock In \emph{Proceedings of the International Conference on Learning Representations (ICLR)}, 2017.
\newblock URL \url{https://openreview.net/forum?id=r1Aab85gg}.

\bibitem[Subramani et~al.(2022)Subramani, Suresh, and Peters]{subramani-etal-2022-extracting}
Nishant Subramani, Nivedita Suresh, and Matthew Peters.
\newblock Extracting latent steering vectors from pretrained language models.
\newblock In \emph{Findings of the Association for Computational Linguistics: ACL}, 2022.
\newblock URL \url{https://aclanthology.org/2022.findings-acl.48}.

\bibitem[Sun(2024)]{jieba_github}
Junyi Sun.
\newblock Jieba: Chinese text segmentation tool.
\newblock Github, 2024.
\newblock URL \url{https://github.com/fxsjy/jieba}.
\newblock Accessed: 2024-09-25.

\bibitem[Tang et~al.(2024{\natexlab{a}})Tang, Yu, Sun, Chen, Tan, Li, Lu, MA, and Zhang]{tang2024salmonn}
Changli Tang, Wenyi Yu, Guangzhi Sun, Xianzhao Chen, Tian Tan, Wei Li, Lu~Lu, Zejun MA, and Chao Zhang.
\newblock {SALMONN}: Towards generic hearing abilities for large language models.
\newblock In \emph{Proceedings of the International Conference on Learning Representations (ICLR)}, 2024{\natexlab{a}}.
\newblock URL \url{https://openreview.net/forum?id=14rn7HpKVk}.

\bibitem[Tang et~al.(2024{\natexlab{b}})Tang, Luo, Huang, Zhang, Wang, Zhao, Wei, and Wen]{tang-etal-2024-language}
Tianyi Tang, Wenyang Luo, Haoyang Huang, Dongdong Zhang, Xiaolei Wang, Xin Zhao, Furu Wei, and Ji-Rong Wen.
\newblock Language-specific neurons: The key to multilingual capabilities in large language models.
\newblock In \emph{Proceedings of the Annual Meeting of the Association for Computational Linguistics (ACL)}, 2024{\natexlab{b}}.
\newblock URL \url{https://aclanthology.org/2024.acl-long.309}.

\bibitem[Templeton et~al.(2024)Templeton, Conerly, Marcus, Lindsey, Bricken, Chen, Pearce, Citro, Ameisen, Jones, Cunningham, Turner, McDougall, MacDiarmid, Freeman, Sumers, Rees, Batson, Jermyn, Carter, Olah, and Henighan]{templeton2024scaling}
Adly Templeton, Tom Conerly, Jonathan Marcus, Jack Lindsey, Trenton Bricken, Brian Chen, Adam Pearce, Craig Citro, Emmanuel Ameisen, Andy Jones, Hoagy Cunningham, Nicholas~L Turner, Callum McDougall, Monte MacDiarmid, C.~Daniel Freeman, Theodore~R. Sumers, Edward Rees, Joshua Batson, Adam Jermyn, Shan Carter, Chris Olah, and Tom Henighan.
\newblock Scaling monosemanticity: Extracting interpretable features from claude 3 sonnet.
\newblock \emph{Transformer Circuits Thread}, 2024.
\newblock URL \url{https://transformer-circuits.pub/2024/scaling-monosemanticity/index.html}.

\bibitem[Tenney et~al.(2019)Tenney, Das, and Pavlick]{tenney-etal-2019-bert}
Ian Tenney, Dipanjan Das, and Ellie Pavlick.
\newblock {BERT} rediscovers the classical {NLP} pipeline.
\newblock In \emph{Proceedings of the Annual Meeting of the Association for Computational Linguistics (ACL)}, 2019.
\newblock URL \url{https://aclanthology.org/P19-1452}.

\bibitem[Todd et~al.(2024)Todd, Li, Sharma, Mueller, Wallace, and Bau]{todd2024function}
Eric Todd, Millicent Li, Arnab~Sen Sharma, Aaron Mueller, Byron~C Wallace, and David Bau.
\newblock Function vectors in large language models.
\newblock In \emph{Proceedings of the International Conference on Learning Representations (ICLR)}, 2024.
\newblock URL \url{https://openreview.net/forum?id=AwyxtyMwaG}.

\bibitem[Toker et~al.(2024)Toker, Orgad, Ventura, Arad, and Belinkov]{toker-etal-2024-diffusion}
Michael Toker, Hadas Orgad, Mor Ventura, Dana Arad, and Yonatan Belinkov.
\newblock Diffusion lens: Interpreting text encoders in text-to-image pipelines.
\newblock In \emph{Proceedings of the Annual Meeting of the Association for Computational Linguistics (ACL)}, 2024.
\newblock URL \url{https://aclanthology.org/2024.acl-long.524}.

\bibitem[Touvron et~al.(2023)Touvron, Martin, Stone, Albert, Almahairi, Babaei, Bashlykov, Batra, Bhargava, Bhosale, et~al.]{touvron2023llama2}
Hugo Touvron, Louis Martin, Kevin Stone, Peter Albert, Amjad Almahairi, Yasmine Babaei, Nikolay Bashlykov, Soumya Batra, Prajjwal Bhargava, Shruti Bhosale, et~al.
\newblock Llama 2: Open foundation and fine-tuned chat models.
\newblock \emph{arXiv preprint}, 2023.
\newblock URL \url{https://arxiv.org/abs/2307.09288}.

\bibitem[Turner et~al.(2024)Turner, Thiergart, Leech, Udell, Vazquez, Mini, and MacDiarmid]{turner2024activationadditionsteeringlanguage}
Alexander~Matt Turner, Lisa Thiergart, Gavin Leech, David Udell, Juan~J. Vazquez, Ulisse Mini, and Monte MacDiarmid.
\newblock Activation addition: Steering language models without optimization.
\newblock \emph{arXiv preprint}, 2024.
\newblock URL \url{https://arxiv.org/abs/2308.10248}.

\bibitem[Vig et~al.(2020)Vig, Gehrmann, Belinkov, Qian, Nevo, Singer, and Shieber]{vig2020investigating}
Jesse Vig, Sebastian Gehrmann, Yonatan Belinkov, Sharon Qian, Daniel Nevo, Yaron Singer, and Stuart Shieber.
\newblock Investigating gender bias in language models using causal mediation analysis.
\newblock In \emph{Advances in Neural Information Processing Systems (NeurIPS)}, 2020.
\newblock URL \url{https://proceedings.neurips.cc/paper_files/paper/2020/file/92650b2e92217715fe312e6fa7b90d82-Paper.pdf}.

\bibitem[Wang et~al.(2024)Wang, Yang, Huang, Yang, Majumder, and Wei]{wang2024multilinguale5textembeddings}
Liang Wang, Nan Yang, Xiaolong Huang, Linjun Yang, Rangan Majumder, and Furu Wei.
\newblock Multilingual e5 text embeddings: A technical report.
\newblock \emph{arXiv preprint}, 2024.
\newblock URL \url{https://arxiv.org/abs/2402.05672}.

\bibitem[Wendler et~al.(2024)Wendler, Veselovsky, Monea, and West]{wendler-etal-2024-llamas}
Chris Wendler, Veniamin Veselovsky, Giovanni Monea, and Robert West.
\newblock Do llamas work in {E}nglish? on the latent language of multilingual transformers.
\newblock In \emph{Proceedings of the Annual Meeting of the Association for Computational Linguistics (ACL)}, 2024.
\newblock URL \url{https://aclanthology.org/2024.acl-long.820}.

\bibitem[Wikimedia-Foundation(2023)]{wikidump}
Wikimedia-Foundation.
\newblock Wikimedia downloads, 2023.
\newblock URL \url{https://dumps.wikimedia.org}.

\bibitem[Wu et~al.(2021)Wu, Peng, and Smith]{sift}
Zhaofeng Wu, Hao Peng, and Noah~A. Smith.
\newblock {Infusing Finetuning with Semantic Dependencies}.
\newblock \emph{Transactions of the Association for Computational Linguistics (TACL)}, 2021.
\newblock URL \url{https://doi.org/10.1162/tacl\_a\_00363}.

\bibitem[Wu et~al.(2024)Wu, Geiger, Arora, Huang, Wang, Goodman, Manning, and Potts]{wu2024pyvenelibraryunderstandingimproving}
Zhengxuan Wu, Atticus Geiger, Aryaman Arora, Jing Huang, Zheng Wang, Noah~D. Goodman, Christopher~D. Manning, and Christopher Potts.
\newblock pyvene: A library for understanding and improving pytorch models via interventions, 2024.
\newblock URL \url{https://arxiv.org/abs/2403.07809}.

\bibitem[Xue et~al.(2021)Xue, Constant, Roberts, Kale, Al-Rfou, Siddhant, Barua, and Raffel]{xue-etal-2021-mt5}
Linting Xue, Noah Constant, Adam Roberts, Mihir Kale, Rami Al-Rfou, Aditya Siddhant, Aditya Barua, and Colin Raffel.
\newblock m{T}5: A massively multilingual pre-trained text-to-text transformer.
\newblock In \emph{Proceedings of the Conference of the North {A}merican Chapter of the Association for Computational Linguistics: Human Language Technologies (NAACL-HLT)}, 2021.
\newblock URL \url{https://aclanthology.org/2021.naacl-main.41}.

\bibitem[Yang et~al.(2023)Yang, Xiao, Wang, Zhang, Bian, Yin, Lv, Pan, Wang, Yan, Yang, Deng, Wang, Liu, Ai, Dong, Zhao, Xu, Sun, Zhang, Liu, Ji, Xie, Dai, Fang, Su, Song, Liu, Ru, Ma, Wang, Liu, Lin, Nie, Guo, Sun, Zhang, Li, Li, Cheng, Chen, Zeng, Wang, Chen, Men, Yu, Pan, Shen, Wang, Li, Jiang, Gao, Zhang, Zhou, and Wu]{yang2023baichuan2openlargescale}
Aiyuan Yang, Bin Xiao, Bingning Wang, Borong Zhang, Ce~Bian, Chao Yin, Chenxu Lv, Da~Pan, Dian Wang, Dong Yan, Fan Yang, Fei Deng, Feng Wang, Feng Liu, Guangwei Ai, Guosheng Dong, Haizhou Zhao, Hang Xu, Haoze Sun, Hongda Zhang, Hui Liu, Jiaming Ji, Jian Xie, JunTao Dai, Kun Fang, Lei Su, Liang Song, Lifeng Liu, Liyun Ru, Luyao Ma, Mang Wang, Mickel Liu, MingAn Lin, Nuolan Nie, Peidong Guo, Ruiyang Sun, Tao Zhang, Tianpeng Li, Tianyu Li, Wei Cheng, Weipeng Chen, Xiangrong Zeng, Xiaochuan Wang, Xiaoxi Chen, Xin Men, Xin Yu, Xuehai Pan, Yanjun Shen, Yiding Wang, Yiyu Li, Youxin Jiang, Yuchen Gao, Yupeng Zhang, Zenan Zhou, and Zhiying Wu.
\newblock Baichuan 2: Open large-scale language models.
\newblock \emph{arXiv Preprint}, 2023.
\newblock URL \url{https://arxiv.org/abs/2309.10305}.

\bibitem[Yang et~al.(2024)Yang, Gribovskaya, Kassner, Geva, and Riedel]{yang-etal-2024-large-language-models}
Sohee Yang, Elena Gribovskaya, Nora Kassner, Mor Geva, and Sebastian Riedel.
\newblock Do large language models latently perform multi-hop reasoning?
\newblock In \emph{Proceedings of the Annual Meeting of the Association for Computational Linguistics (ACL)}, 2024.
\newblock URL \url{https://aclanthology.org/2024.acl-long.550}.

\bibitem[Zeng et~al.(2024)Zeng, Han, Chen, and Yu]{zeng2024converginglinguafrancaevolution}
Hongchuan Zeng, Senyu Han, Lu~Chen, and Kai Yu.
\newblock Converging to a lingua franca: Evolution of linguistic regions and semantics alignment in multilingual large language models, 2024.
\newblock URL \url{https://arxiv.org/abs/2410.11718}.

\bibitem[Zhao et~al.(2024)Zhao, Zhang, Chen, Kawaguchi, and Bing]{zhao2024largelanguagemodelshandle}
Yiran Zhao, Wenxuan Zhang, Guizhen Chen, Kenji Kawaguchi, and Lidong Bing.
\newblock How do large language models handle multilingualism?
\newblock \emph{arXiv preprint}, 2024.
\newblock URL \url{https://arxiv.org/abs/2402.18815}.

\end{thebibliography}
\bibliographystyle{iclr2024_conference}

\appendix

\section{Experimental Details for \S\ref{sec:identify}}

We first note a tokenization detail regarding ``prefix spaces.'' For example, \token{\_llama} (where \_ denotes whitespace) is a token in the Llama-3 vocabulary, but not \token{llama}. When comparing token-level logit lens probabilities, we use the configuration that maximizes the probability, which may be different for each setting. For example, the token with the prefix space usually is more likely, but for arithmetic expressions \token{a=b+c}, after \token{+}, the surface token \token{c} is more likely than \token{\_c}.

\subsection{Multilingual} \label{sec:appendix-multilingual}

For Experiment 1, for each sentence pair, we use a template to transform each sentence. This is due to the automatic code-switching behavior of LMs. For an English model processing Chinese text, we expect the Chinese tokens to have high probabilities in the final layer because they need to be output; however, we observe these models tend to code-switch back to their dominant language after a full sentence, which confounds our analysis. We therefore put the parallel sentences into a template, ``\{English Sentence\} This represents'' (and the corresponding Chinese version), as the model is less likely to code-switch mid-sentence after ``represents''. We experimented with other templates that led to similar results. Furthermore, for sentence in GALE~\citep{gale2016}, we make sure the \texttt{transcript;unicode} is not empty for both the source and the translation.

For Experiment 2, due to tokenization, it is challenging to obtain exactly parallel English-Chinese tokens, and hence we perform aggressive filtering. We consider only text positions where the next BPE token (1) is a valid Chinese word (as segmented by Jieba~\citep{jieba_github}), and (2) has an English translation (using the English-Chinese dictionary CC-CEDICT~\citep{mdbg_cc_cedict}). E.g., \chin{``今天是开心的一天''} (Today is a happy day), Llama-3 tokenizes it as [\chin{`今天', `是', `开', `心', `的一', `天'}], while Jieba segments it as [\chin{`今天', `是', `开心', `的', `一天'}]. We keep \{\chin{`今天', `是'}\}. Furthermore, only \chin{`今天'}'s translation is a single token, the only token that survives the cutoff is \chin{`今天'}.

\subsection{Code} \label{sec:appendix-identify-code}

For all experiments using the MBPP dataset, we concatenate all dataset splits with a total of 974 programs. For the function call argument experiment, we consider all non-zero-argument function calls in MBPP, excluding unit tests. We automatically identify the argument names (the ``semantic roles'') by function inspection for built-in functions and by looking at the function definition for those defined in-context, and skip when this is not possible. We also ignore arguments whose semantic roles are generically called \token{obj} or \token{object}, and instances where the instantiated surface-form argument is the same as the semantic role. We look the hidden state corresponding to the previous token, either \token{(} or \token{,}, except when tokenization renders this impossible (e.g., when the previous token is merged with a part of the surface argument; we find this happens often with the Llama-3 tokenizer and thus do not include it in this experiment). This leaves 540 arguments.

\subsection{Vision-Language} \label{sec:appendix-identify-vision-language}

To pass the images through the model, we embed them in templates, only for the logit lens experiments. For the color experiment, we use the template
\token{USER: What is the color in the image?<image>\textbackslash n ASSISTANT:}. For the caption and segmentation experiments, we use
\token{USER: What is in the image?\textbackslash n<image> ASSISTANT:}
for LLaVA and 
\token{What is in the image?\textbackslash n<image>}
for Chameleon.

For all caption and segmentation experiments, we use the MSCOCO 2017 dataset.
In particular, for the segmentation evaluation, we use the MSCOCO 2017 panoptic segmentation labels. We evaluate the alignment score between all corresponding patches and labels. We consider a patch and a label as ``corresponding'' if there is an image segment with that label that occupies more than half of the pixels in the patch.

\section{Experimental Details for \S\ref{sec:intervention}} \label{sec:appendix-intervention}

For both the code and vision-language intervention experiments, we use argmax decoding. We perform most of our intervention experiments using the \texttt{pyvene} library~\citep{wu2024pyvenelibraryunderstandingimproving}.

\subsection{Multilingual} \label{sec:appendix-intervention-multilingual}

For each language, we sample $N=1000$ instances from the training set of InterTASS for Spanish and the multilingual Amazon reviews corpus for Chinese.
Following \citet{turner2024activationadditionsteeringlanguage}, we use trained models for various metrics.
We found that Llama models tend to code-switch back to English when processing texts in other languages, and discovered that we can mitigate this with an instruction: ``\chin{接下来的文字全部是中文的。}'' for Chinese and ``Todo el texto siguiente está en español.'' for Spanish (trans. ``All of the following text is in Chinese/Spanish.'').

We automatically evaluate the sentiment of the generation using a DistillBERT-based~\citep{sanh2020distilbertdistilledversionbert} model finetuned for multilingual sentiment analysis,\footnote{\url{https://huggingface.co/lxyuan/distilbert-base-multilingual-cased-sentiments-student}} judge the generation fluency by taking the conditional perplexity of the generation given the prefix from Llama-3.1-70B~\citep{dubey2024llama3herdmodels},\footnote{We also tried using Mistral-Nemo-Base (\url{https://huggingface.co/mistralai/Mistral-Nemo-Base-2407}), and found similar trends.} and compute the relevance of the generation with the prefix by computing the cosine similarity between the generation and the prefix using a XLM-R-Large-based~\citep{conneau-etal-2020-unsupervised} model finetuned for sentence representation\footnote{\url{https://huggingface.co/intfloat/multilingual-e5-large}}~\citep{wang2024multilinguale5textembeddings}. All these models support both Spanish and Chinese.

We perform ActAdd by passing both the positive and negative steering words through the LM, taking their hidden states at layer 17, computing their difference, scaling it by a constant, and adding it to the normal generation forward pass also at layer 17, exactly following \citet{turner2024activationadditionsteeringlanguage}, except we use a scaling coefficient of 5, rather than 2 in their experiments, for which we observed a larger effect.
For generation, we use a temperature of 1, top-$p$=0.3, and a frequency penalty of 1, all following \citet{turner2024activationadditionsteeringlanguage}, without tuning.

We showed the Llama-3 intervention results in Table~\ref{tab:actadd}, and here in Table~\ref{tab:actadd-llama2} we show the results on Llama2, with similar trends.

\begin{table*}[t!]
    \centering
    \caption{\label{tab:actadd-llama2}
    Steering Llama-2's output sentiments using trigger words in English vs. the input language (either Spanish or Chinese). We report the mean sentiment, disfluency (perplexity), and relevance of the continuation, as well as the standard deviation across 10 seeds.  \textbf{Cross-lingual steering is consistently successful, sometimes even more than monolingual steering, without substantial damage in text fluency and relevance.}
    }
    \footnotesize
    \begin{tabular}{ccc|ccc}
        \toprule
        Text Lang. & Steering Dir. & Steering Lang. & Sentiment & Disfluency ($\downarrow$) & Relevance ($\uparrow$) \\
        \midrule
        \multirow{5}[3]{*}{Spanish} & None & None & 0.144$_{\pm0.014}$ & \phantom{00}8.58$_{\pm0.57}$ & 0.850$_{\pm0.006}$ \\
        \cmidrule{2-6}
        & \multirow{2}{*}{$\downarrow$} & Spanish & 0.143$_{\pm0.012}$ & \phantom{00}8.84$_{\pm0.79}$ & 0.847$_{\pm0.006}$ \\
        && English & 0.097$_{\pm0.024}$ & \phantom{00}8.99$_{\pm0.72}$ & 0.847$_{\pm0.005}$ \\
        \cmidrule{2-6}
        & \multirow{2}{*}{$\uparrow$} & Spanish & 0.164$_{\pm0.018}$ & \phantom{00}9.11$_{\pm0.50}$ & 0.844$_{\pm0.005}$ \\
        && English & 0.149$_{\pm0.015}$ & \phantom{00}8.35$_{\pm0.30}$ & 0.849$_{\pm0.006}$ \\
        \cmidrule{1-6}
         \multirow{5}[3]{*}{Chinese} & None & None & 0.223$_{\pm0.036}$ & \phantom{0}14.63$_{\pm2.65}$ & 0.844$_{\pm0.009}$ \\
        \cmidrule{2-6}
        & \multirow{2}{*}{$\downarrow$} & Chinese & 0.117$_{\pm0.080}$ & \phantom{0}15.29$_{\pm2.47}$ & 0.840$_{\pm0.011}$ \\
        && English & 0.156$_{\pm0.076}$ & \phantom{0}14.80$_{\pm2.24}$ & 0.842$_{\pm0.008}$ \\
        \cmidrule{2-6}
        & \multirow{2}{*}{$\uparrow$} & Chinese & 0.359$_{\pm0.077}$ & 545.94$_{\pm1544.36}$ & 0.839$_{\pm0.010}$ \\
        && English & 0.227$_{\pm0.038}$ & \phantom{0}14.14$_{\pm2.42}$ & 0.845$_{\pm0.009}$ \\
        \bottomrule        
    \end{tabular}
\end{table*}

\section{Token-Level Multimodal Experiments Using the Logit Lens} \label{sec:appendix-logitlens}

As mentioned in \S\ref{sec:method-new}, we do not perform a logit lens analysis for the multimodal experiments. But we can perform a logit-lens-style test for Eq.~\ref{eq:rep-sim-enc} that test representation equivalence not on a sequence level but on a token level. At each individual token (e.g., image patches), we measure the alignment of it representation and a natural language description token of it, using the logit lens.

\begin{figure}[t!]
    \begin{subfigure}[t]{0.242\textwidth}
        \centering
        \includegraphics[width=\textwidth]{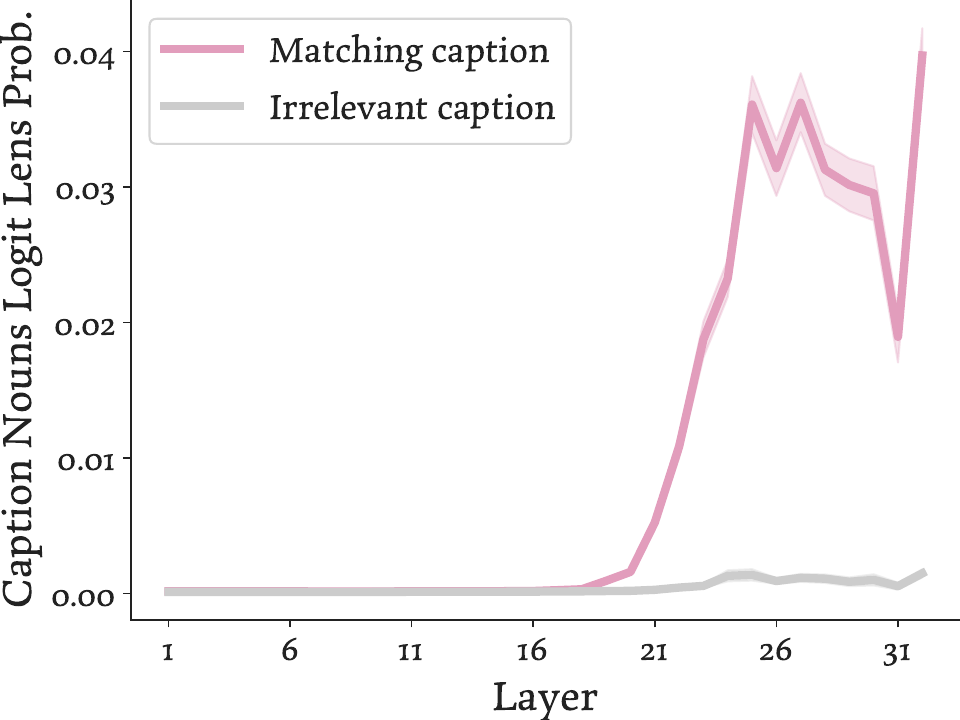}
        \caption{Caption, LLaVA}
        \label{fig:caption-llava}
    \end{subfigure}%
    ~
    \begin{subfigure}[t]{0.242\textwidth}
        \centering
        \includegraphics[width=\textwidth]{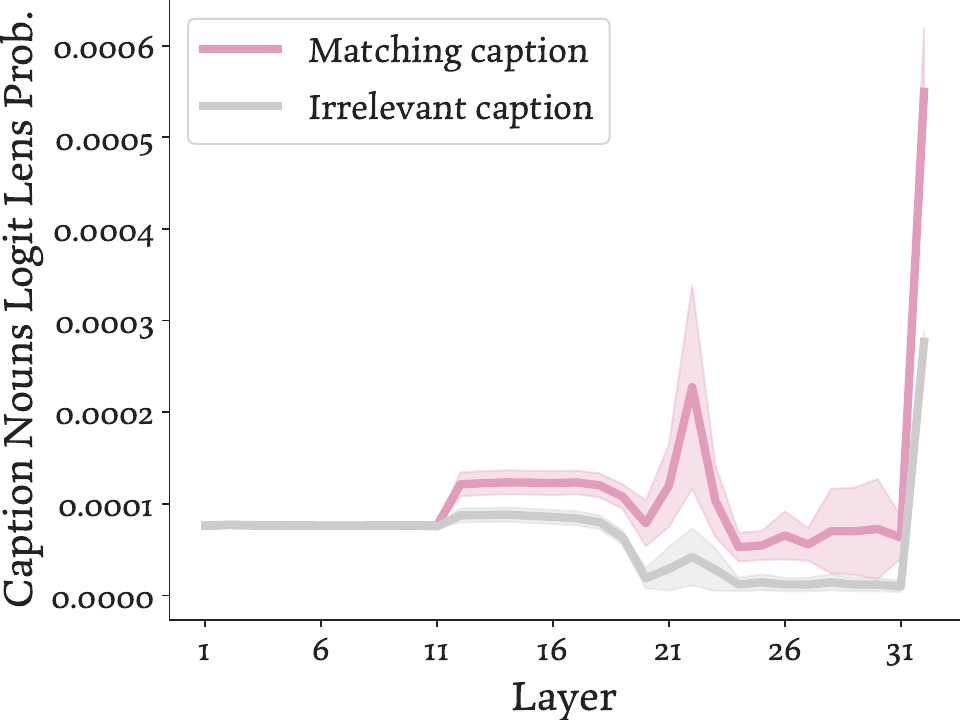}
        \caption{Caption, Chameleon}
        \label{fig:caption-chameleon}
    \end{subfigure}%
    ~
    \begin{subfigure}[t]{0.242\textwidth}
        \centering
        \includegraphics[width=\textwidth]{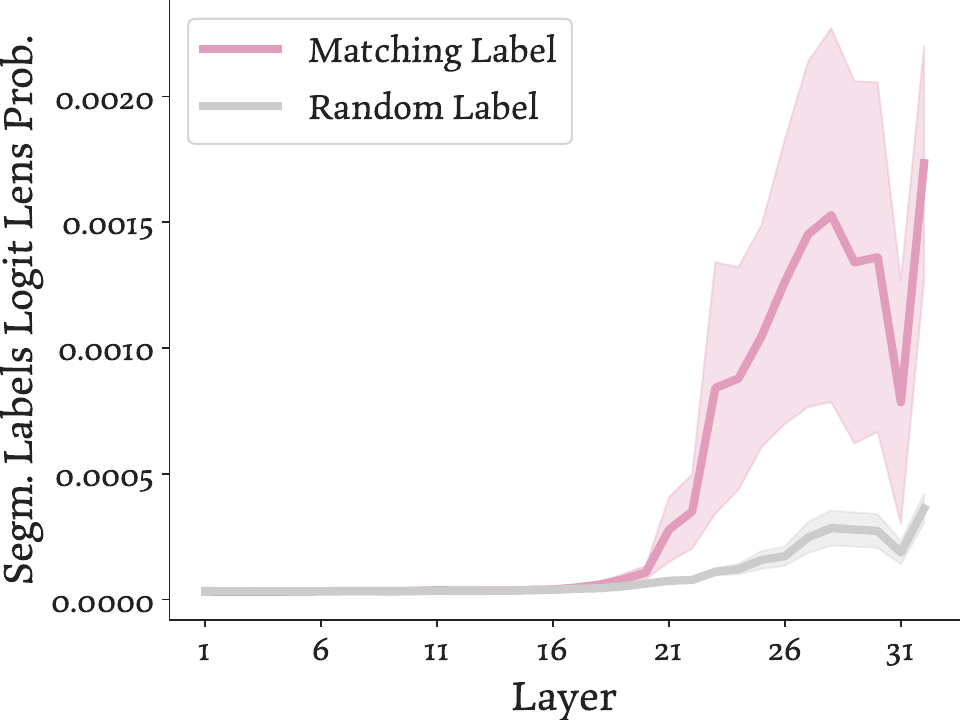}
        \caption{Segment., LLaVA}
        \label{fig:segmentation-llava}
    \end{subfigure}%
    ~
    \begin{subfigure}[t]{0.242\textwidth}
        \centering
        \includegraphics[width=\textwidth]{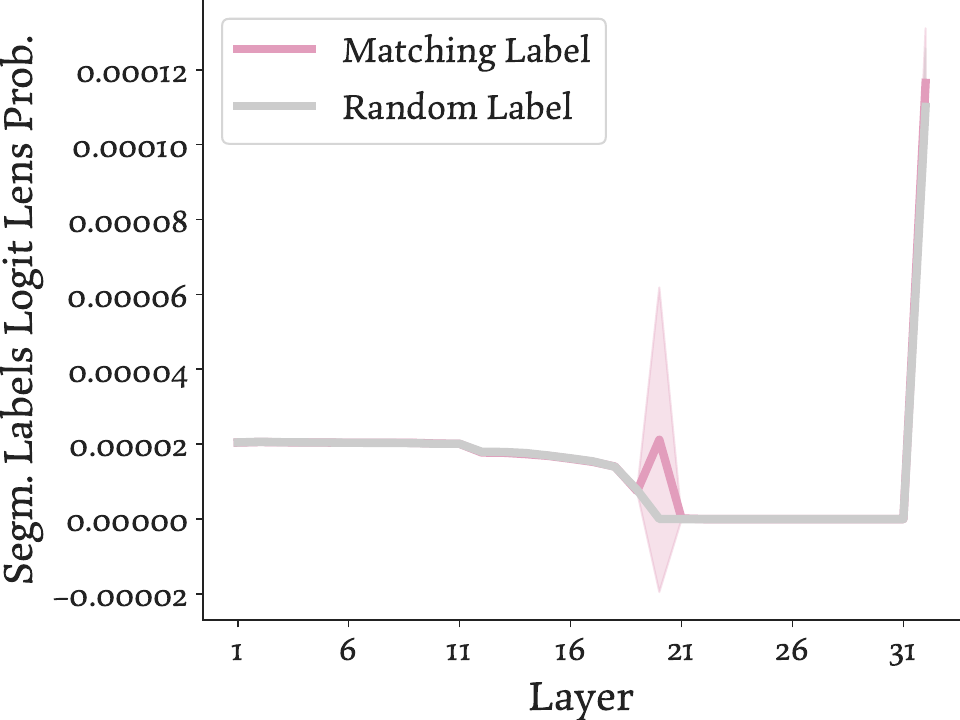}
        \caption{Segment., Chameleon}
        \label{fig:segmentation-chameleon}
    \end{subfigure}
    \caption{When processing an image patch, model logit lens probabilities of either the nouns in the corresponding caption or the patch segmentation label, as well as a baseline for each with no correspondence between the patch and the label. \textbf{The image representations better match the semantically corresponding English words.}}
\end{figure}

\subsection{Visual Input.} \label{sec:logit-lens-multimodal}

We consider the alignment of image tokens (corresponding to patches in the original image) to their description words. First, we use the nouns\footnote{Words with \texttt{NOUN} or \texttt{PROPN} tags given by SpaCy's \texttt{en\_core\_web\_trf} model~\citep{spacy2}.} in the image caption as a coarse description---although they do not precisely describe all patches, they should better align with the patches on average than nouns irrelevant to the image. We consider the same 1000 image captions as in \S\ref{sec:identify-visionlanguage}. For each patch, we compute a scalar patch-caption alignment score (for each layer separately) by summing over the logit lens probabilities for all the nouns in the caption at that image patch position.
We average this alignment score over all patches in all images, separately for each transformer layer. For the irrelevant nouns baseline, for each image, we compute the alignment score with an unrelated caption that has the smallest noun overlap with the groundtruth matching caption (we also normalize the number of nouns so that the score is comparable). Figures~\ref{fig:caption-llava} and \ref{fig:caption-chameleon} show that the matching caption better aligns with the image patch representations than an unmatched caption for both LLaVA and Chameleon.

Image segmentation labels, which annotate objects in specific image locations, provide a finer-grained patch description.
The setup is near-identical to captions, but the scalar alignment score for a given patch is not computed using its logit lens probability of caption nouns, but of its corresponding object label.
We say a label correspond to the patch if more than half of the patch's pixels have that label.
For the irrelevant token baseline, we compute the alignment by aligning each patch with a different randomly chosen object category from all categories. Figures~\ref{fig:segmentation-llava} and \ref{fig:segmentation-chameleon} show that, for LLaVA, the patches are much better aligned to the corresponding labels than randomly assigned labels (which have near-0 logit lens probability).
For Chameleon, this is the case for only one middle layer, and not in a statistically significant way, though, as we showed in \S\ref{sec:intervention-visionlanguage}, Chameleon's latent space can be reliably steered using English tokens.

\subsection{Audio} \label{sec:logit-lens-audio}

\begin{wrapfigure}{R}{4.5cm}
\vspace{-4.5pt}
    \begin{minipage}{0.3\textwidth}
        \centering
    \includegraphics[width=\textwidth]{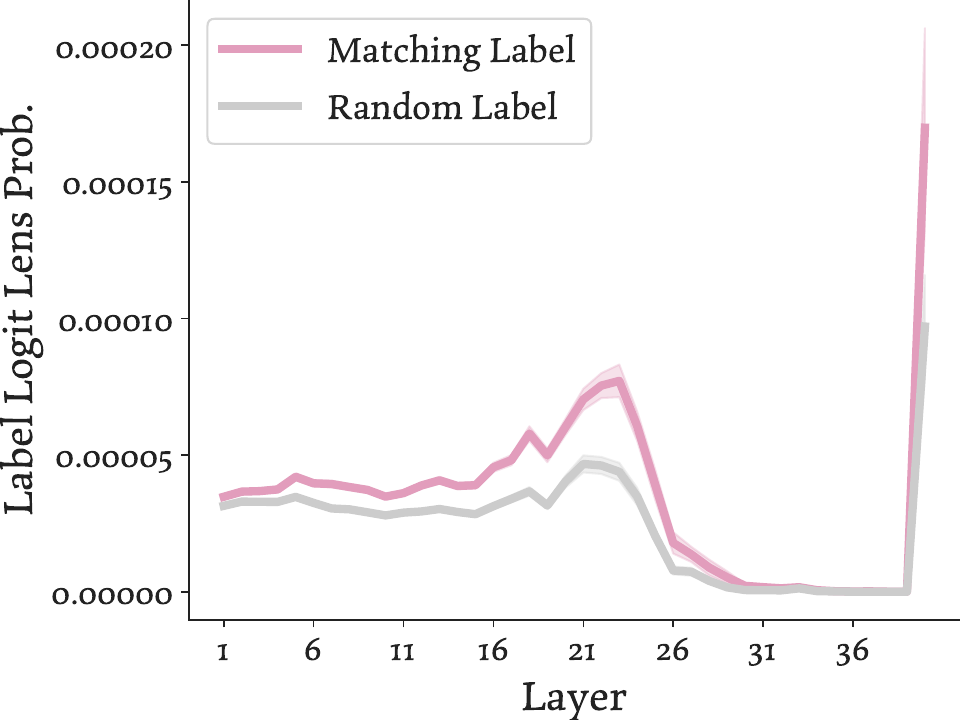}
    \caption{When SALMONN processes an audio clip, the logit lens probabilities of the English words in the audio label vs. another random label. \textbf{The audio representations better match the semantically corresponding label in English.}}
    \label{fig:audio}
    \end{minipage}
    \vspace{-13pt}
\end{wrapfigure}

Unlike for vision-language models where we can map individual image patches to model input token positions, such correspondence does not exist in SALMONN. This limits us to position-agnostic evaluations like the captioning study, preventing a fine-grained analysis such as using segmentation labels. 
Similar to the captioning experimental design, we measure the average logit lens probabilities of the words in the label, and consider a random label in the dataset with no word overlap as the baseline. %
On the same 1000 samples, Figure~\ref{fig:audio} shows a familiar trend, where the audio hidden states are closer to semantically corresponding label words. We note that this is a lower bound---many words in some labels, such as the prepositions in the label \token{writing on blackboard with chalk}, are unlikely to be represented in the audio hidden states.

\end{document}